\documentclass[review]{elsarticle}

\usepackage{amsmath,amssymb,amsfonts}
\usepackage{textcomp}
\usepackage{bm}
\usepackage{array}
\usepackage{booktabs}
\usepackage{soul}
\usepackage{changes}
\definechangesauthor[name={Per cusse}, color=orange]{per}
\usepackage{algorithm}
\usepackage{algorithmic}
\newcommand{\norm}[1]{\left\lVert#1\right\rVert}

\journal{Journal of \LaTeX\ Templates}

\begin{document}

\begin{frontmatter}

\title{Construct Informative Triplet with Two-stage Hard-sample Generation}


\author[mymainaddress]{Chuang Zhu\fnref{fn1}\corref{cor1}}
\ead{czhu@bupt.edu.cn}
\author[mymainaddress]{Zheng Hu\fnref{fn1}}
\author[mymainaddress]{Huihui Dong}
\author[mymainaddress]{Gang He}
\author[mysecondaryaddress]{Zekuan Yu\corref{cor1}}
\author[mythridaddress]{Shangshang Zhang}
\fntext[fn1]{The first two authors contributed equally to this work.}

\cortext[cor1]{Corresponding author}


\address[mymainaddress]{School of Information and Communication Engineering, Beijing University of Posts and Telecommunications, Beijing, China}
\address[mysecondaryaddress]{Center for Shanghai Intelligent Imaging for Critical Brain Diseases Engineering and Technology Research, Fudan University, Shanghai, China}
\address[mythridaddress]{Berkeley AI Research Lab, University of California Berkeley, CA, 94720, USA}

\begin{abstract}
In this paper, we propose a robust sample generation scheme to construct informative triplets. The proposed hard sample generation is a two-stage synthesis framework that produces hard samples through effective positive and negative sample generators in two stages, respectively. The first stage stretches the anchor-positive pairs with piecewise linear manipulation and enhances the quality of generated samples by skillfully designing a conditional generative adversarial network to lower the risk of mode collapse. The second stage utilizes an adaptive reverse metric constraint to generate the final hard samples. Extensive experiments on several benchmark datasets verify that our method achieves superior performance than the existing hard-sample generation algorithms. Besides, we also find that our proposed hard sample generation method combining the existing triplet mining strategies can further boost the deep metric learning performance.
\end{abstract}

\begin{keyword}
{Deep metric learning}\sep triplet \sep hard sample generation \sep adversarial network \sep  mode collapse.
\end{keyword}

\end{frontmatter}


\section{Introduction}
Many applications, such as mobile augmented reality (MAR) and social life based on image websites, are becoming more and more popular with the increase of digital cameras and intelligent mobile devices \cite{li2015weakly}. Millions of images are produced every day from these types of digital equipment, and thus how to accurately search images from a large dataset poses a great challenge. To address this problem, the content-based image retrieval (CBIR) technique is proposed to search for images representing the same object or scene as the one depicted in a query image \cite{girod2011mobile,zhu2019feature}. 

The key for CBIR is how to build compact and robust image representations. The traditional hand-crafted features suffer large memory size consumption and will lower the searching efficiency \cite{girod2011mobile,lowe2004distinctive}. Many compact image representations \cite{perronnin2010large,jegou2011aggregating,azizpour2015generic,babenko2015aggregating,tolias2015particular} are thus proposed to address the above problem. Among these studies, the deep features can produce better performance than the traditional Fisher compact features (such as work \cite{perronnin2010large} and \cite{jegou2011aggregating}), and they have become the predominant stream of features used for CBIR. Generally, the Convolutional Neural Networks (CNN) layer activations are directly adopted as the off-the-shelf deep features \cite{azizpour2015generic,babenko2015aggregating,tolias2015particular}. 
However, these features are usually trained for image classification tasks. They should be further optimized by transferring the CNN model to the image retrieval task through the deep metric learning (DML) architecture \cite{arandjelovic2016netvlad,radenovic2016cnn}.

In deep metric learning, the matching/non-matching training pairs or triplets are first carefully constructed and then passed to the Siamese \cite{hu2014discriminative} or triplet architectures \cite{hoffer2015deep}. The off-the-shelf deep feature is thus fine-tuned by decreasing the distances of matching pairs and increasing the distances of non-matching pairs. How to construct the informative matching/non-matching pairs or triplets plays a vital role in the fine-tuning process. Typically, there are two different lines of recent research regarding this issue: hard example mining and hard example generation. 

In the first category, the researchers applied hard example mining to build the pairs or triplets, which targets selecting samples that can provide informative supervision to make the model training more efficient \cite{wu2017sampling,yuan2017hard,radenovic2018revisiting,radenovic2018fine}. {The hard mining strategy has been proved to be effective in accelerating the convergence of the network. However, this training strategy may lead to training a biased network \cite{yu2018correcting} because only a few samples are selected for training, while most of the non-selected easy samples are considered invalid. }

\begin{figure}[h]
  \centering
  \includegraphics[width=0.95\linewidth]{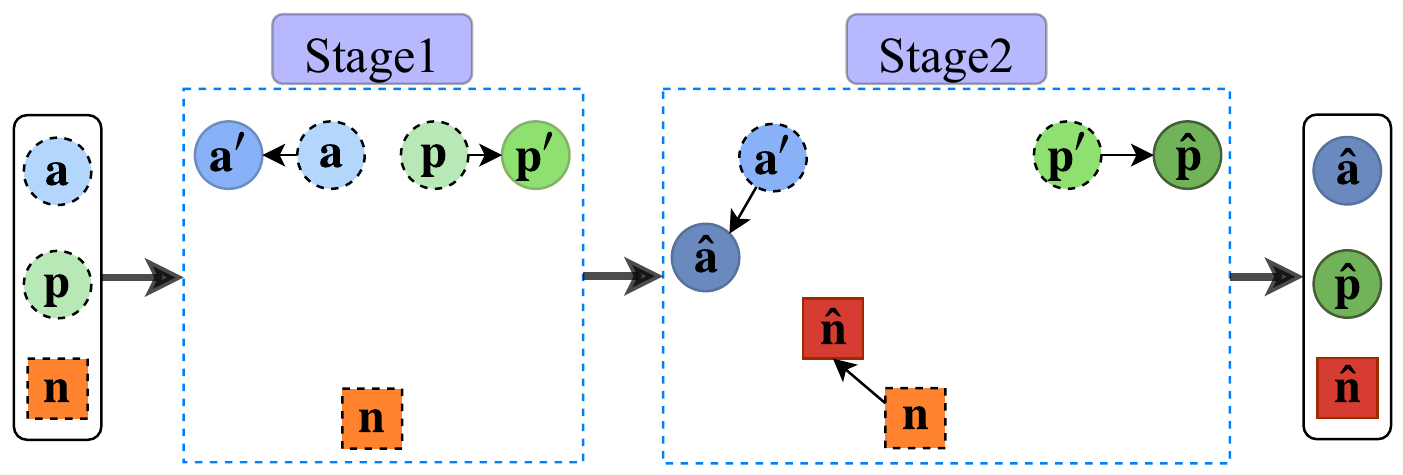}
  \caption{{Schematic diagram of our proposed hard sample generation. In Stage 1, anchor and positive embeddings $\{\mathbf{a},\mathbf{p}\}$ will be pushed away from each other; in Stage 2, the hard negative embedding will be generated with reversed metric loss, and the hard triplet is thus produced by adjusting the anchor-positive pair $\{\mathbf{a}^\prime,\mathbf{p}^\prime\}$. The generated hard triplet \{$\hat{\mathbf{a}}$, $\hat{\mathbf{p}}$, $\hat{\mathbf{n}}$}\} is thus applied to deep representation learning. }  
  \label{fig:trip-gen}
\end{figure}

The second category tries to construct informative triplets through hard example generation. These studies focus on digging the potential information of easy samples by generating hard samples with one-stage adversarial model, instead of just mining existing samples \cite{zhao2018adversarial,duan2018deep}. 
This kind of research utilizes both the original and the generated hard samples to construct the informative pairs or triplets, and then optimizes the embedding learning. However, the positive and negative samples play different roles in the deep metric learning process, which have different distributions and characteristics. It is difficult to exploit the full potential capability of all the positive and negative samples at the same time just through one adversarial model. 







\textbf{Approach and Contributions.}
{In this work, we propose a novel hard sample generation for deep embedding learning.  {To exploit the full potential ability of all the simple samples, we propose a two-stage generator to perform the hard positive and negative generation, separately. Moreover, to avoid generating random vectors, the existing hard sample generation schemes adopt label preserving synthesis, which just requires the generated samples to have the same label with their original samples. However, the generator can only produce samples from a subset of modes to meet the above simple constraint. This will ruin the diversity of the generated hard samples and thus cause the mode collapse problem \cite{guo2019mode}. We propose to build more powerful conditional synthesis with stronger constraints to alleviate this problem.}} The main contributions of our work are summarized as follows: 
\begin{itemize}

\item A two-stage adversarial learning architecture, as shown in Fig \ref{fig:trip-gen}, is proposed to produce hard positives and hard negatives at different stages, respectively. In this architecture, the hard anchor-positive pair is first produced and then the hard negative generation is performed in the second stage. Through our two-stage generation, the potential ability of both the simple positives and negatives is mined.

\item In the hard anchor-positive generation, we perform a stronger conditional synthesis scheme than the recent label-preserving method. {Different from the previous works, we use a conditional generative adversarial network that introduces intermediate embeddings as discriminatory conditions to generate samples that can be as close as the original samples and lower the risk of mode collapse. Besides, a piecewise linear manipulation function is also proposed to properly control the hard level of the generated samples.} 

\item In the hard negative generation, we propose an adaptive reverse triplet loss (ART-loss). We impose more stringent restrictions on reverse triplet loss with a larger threshold when the sample generation is getting better. With ART-loss we can gradually generate harder and harder negative samples. 

\item {We experimentally demonstrate that our proposed two-stage hard sample generation can be directly combined with other triplet mining methods. Our method can generate informative hard samples, which can be used as a complement to previous hard mining strategies to further boost the DML model performance.}
\end{itemize}

\textbf{Outline.} {The paper is organized as follows. In Section \ref{sec:RelatedWork}, we review techniques that are related to our work. In Section \ref{sec:pre}, we show some preliminaries about deep metric learning. In Section \ref{sec:framework}, we build the whole deep embedding learning framework. After that, we present the proposed two-stage hard sample generation scheme in Section \ref{sec:hsg}. Then, we report the experimental results in Section \ref{sec:experiment}. Finally, the conclusions of this paper are summarized in Section \ref{sec:conclusion}.}

\section{Related Works}
\label{sec:RelatedWork}
\textbf{Hard Sample Mining.} {Because not all image pairs or triplets are equally informative, many works proposed schemes mining hard samples to train the deep metric models \cite{wu2017sampling, yuan2017hard, cui2016fine, hermans2017defense, Yu_2018_ECCV}.} Hard sample mining is a technique for selecting informative sample pairs in deep metric learning. Considering that the number of triplets grows cubically with the size of training data, triplet selection is quite necessary for training with triplet loss. These triplets can be selected not only offline from the entire training set \cite{hermans2017defense}, but also from each batch of training using an online approach  \cite{wu2017sampling, Yu_2018_ECCV, schroff2015facenet}. Hard negative pair mining \cite{hermans2017defense} proposed the selection of the hardest positive and negative within a batch to construct triplets that produce useful gradients and therefore help triplet loss networks converge quickly. While it speeds up convergence, it also could lead to a bad local minimum and a biased model due to the hardest positive and negative may often be noise in data. In work \cite{schroff2015facenet}, the authors proposed a more relaxed strategy, semi-hard negative pair mining, which chooses an anchor-negative pair that is farther than the anchor-positive pair, but within a margin. Based on this idea, soft-hard mining \cite{Yu_2018_ECCV} restricting the sampling set using moderate negatives and positives {is proposed to avoid too confusing samples that are highly likely to be noisy data}. Given that the previous method only focuses on a small number of hard triplets and ignores a large number of simple samples, distance-weighted tuple mining \cite{wu2017sampling} comprehensively considers samples in the whole spectrum of difficulties by introducing a sampling distribution over the range of distances between anchor and negatives.

\textbf{Hard Sample Generation.} The more recent researches seek approaches that can generate hard sample pairs or triplets to optimize the deep network, such as work \cite{zhao2018adversarial}, work \cite{duan2018deep}, work \cite{zheng2019hardness} and \cite{gu2020symmetrical}.


In work \cite{zhao2018adversarial}, the authors proposed an adversarial learning algorithm to jointly optimize a hard triplet generator and an embedding network. The hard triplet generation is realized by using an inverse triplet loss function, and the generator is constrained by keeping label consistency to avoid random output. However, in metric learning, just the generated hard triplets are adopted and the original triplets are ignored. Besides, performing hard triplet generation only through one adversarial stage fails to dig the full potential information for both the simple positive and negative samples at the same time. Both work \cite{duan2018deep} and \cite{zheng2019hardness} focus on hard negative generation. Compared with work \cite{duan2018deep}, the authors in \cite{zheng2019hardness} proposed an adaptive hardness-aware augmentation method to control the hard levels of the generated samples and thus achieved some improvement. The hard level here denotes the ``hard degree" of the image matching pair or non-matching pair used in the deep metric learning: the hard level is high when the embedding distance is big for the matching pair or the embedding distance is small for the non-matching pair, and vice versa. However, these two studies just digging the potential information of easy negatives while ignoring the easy positives. All the above hard sample generation studies just use the simple label preserving technique to avoid the failure of generation. In work \cite{gu2020symmetrical}, the authors proposed a method of synthetic hard sample generation, which generates symmetrical synthetic embeddings with each other as an axis of symmetry and then selects the hardest negative pair within the original and synthetic embeddings. Although this scheme is computationally inexpensive, it can only synthesise samples at corresponding symmetric positions in the embedding space, and the diversity of the synthesised samples is very dependent on the number of original samples.

\section{Preliminaries}
\label{sec:pre}

In this paper, the boldface uppercase letter denotes a set (pair) of input images, embeddings, or labels; the boldface lowercase letter or the uppercase letter denotes a specific image, embedding, or label. 
The main notations used in this paper are listed in Table \ref{tab:notations}.

\begin{table}[t]
	\centering
	\label{tab:notations}
\begin{tabular}{l|l}
\hline
Notations & Descriptions \\ \hline
$\mathbf{I}$   & the input images
\\ \hline
$\mathbf{L}$, $\hat{\mathbf{L}}$ & the ground truth labels and predicted labels \\ \hline
${\mathbf{X}}$, $\hat{\mathbf{X}}$ & original samples and the generated hard samples  \\ \hline
$\mathbf{a}$, $\mathbf{p}$, $\mathbf{n}$ & a triplet embeddings of the anchor, positive and negative \\ \hline
$\hat{\mathbf{a}}$, $\hat{\mathbf{p}}$, $\hat{\mathbf{n}}$ & generated hard triplet embeddings \\ \hline
$I_a$,$I_p$,$I_n$ & a triplet input images \\ \hline
$l$, $\hat{l}$ & a ground truth label and a predicted label 
\\ \hline
\end{tabular}
	\caption{Main notations and their descriptions.}
\end{table}
{Deep metric learning aims to learn an embedding that can decrease the distance between matching pairs and increase the distance between non-matching pairs \cite{zhu2019feature,sohn2016improved}. To learn such an embedding, many works are proposed, such as the traditional contrastive loss \cite{hu2014discriminative,chopra2005learning} under Siamese network and triplet loss \cite{hoffer2015deep,schroff2015facenet}.
}

\begin{figure}[h]
  \centering
  \includegraphics[width=2.78in]{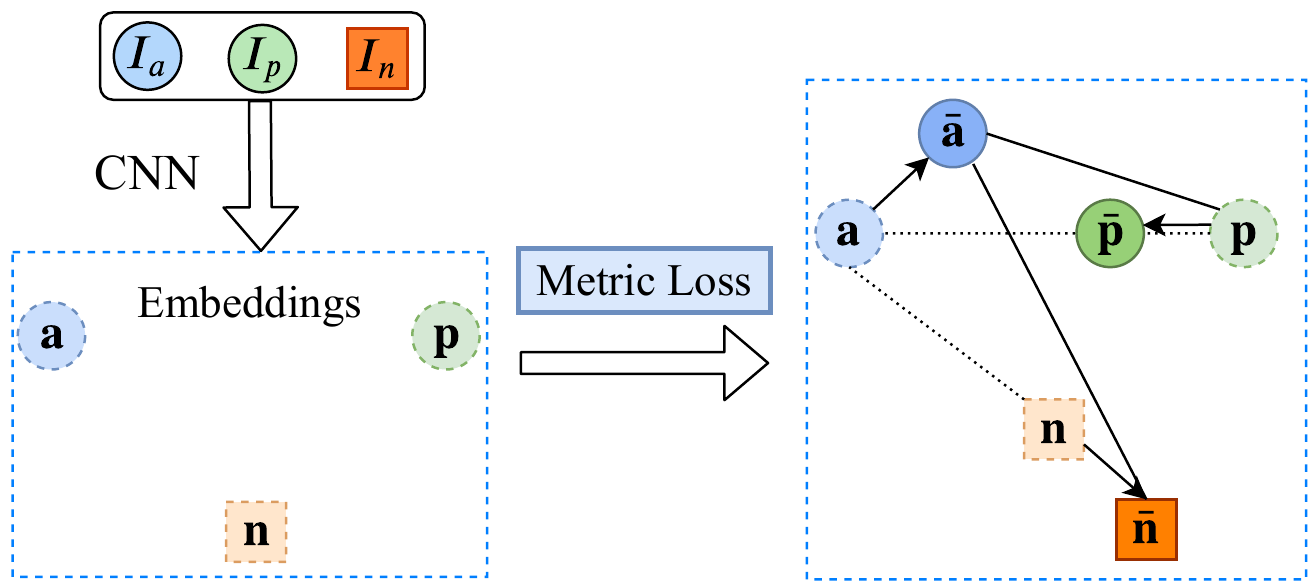}
  \caption{{Triplet loss architecture. After metric learning, the distance between anchor $\mathbf{a}$ and positive embedding $\mathbf{p}$ becomes smaller, and the negative embedding $\mathbf{n}$ is pushed away from the anchor embedding.}}
  \label{fig:pre_triplet}
\end{figure}

{In this paper, we apply the triplet architecture to learn the embedding for deep metric learning.}
Triplet loss-based metric learning takes a triplet as the input, where each triplet contains an anchor (${{I}_a}$), a positive (${I}_p$), and a negative (${I}_n$) image. The triplet loss is written as:
\begin{align}
\mathcal{L}_\text{t}&=\left[ \norm{F({I}_a)-F({I}_p)}^2_2-\norm{F({I}_a)-F({I}_n)}^2_2+\tau  \right]_+ \notag\\ 
&=\left[ \norm{\mathbf{a}-\mathbf{p}}^2_2-\norm{\mathbf{a}-\mathbf{n}}^2_2+\tau  \right]_+
\label{eq:trip}
\end{align}
where $\{\mathbf{a},\mathbf{p},\mathbf{n}\}$ are the learned triplet embeddings corresponding to $\{I_{a},I_{p},I_{n}\}$; $F(\cdot)$ denote the learned deep network which maps the input image to the embedding space; the operator $\left[\cdot\right]_+$ is the max(0, $\cdot$) function, and $\tau>0$ is a distance margin that makes the anchor-positive pair closer than the anchor-negative pair. The triplet based metric learning is depicted in Fig. \ref{fig:pre_triplet}.  

\section{Deep Embedding Learning Framework with Hard-sample Generation}
\label{sec:framework}
\subsection{Framework}

\begin{figure*}[hbtp]
  \centering
  \includegraphics[width=0.95\linewidth]{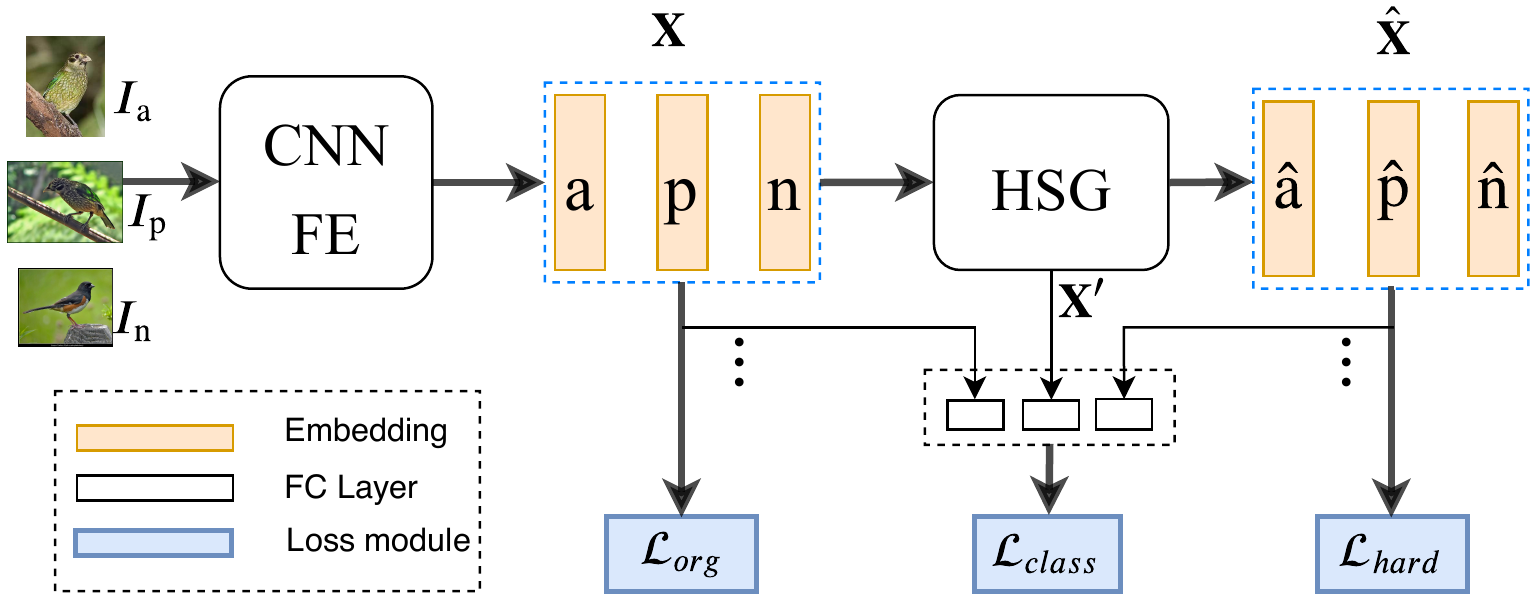}
  \caption{The proposed deep embedding learning framework. The original samples $\mathbf{X} = \{\mathbf{a}, \mathbf{p}, \mathbf{n}\}$ are used to perform the original triplet loss $\mathcal{L}_{org}$; the generated hard samples produced by HSG $\hat{\mathbf{X}} = \{\hat{\mathbf{a}}, \hat{\mathbf{p}}, \hat{\mathbf{n}}\}$ are applied to yield the generated triplet loss $\mathcal{L}_{hard}$; all the categories of original samples $\mathbf{X}$, temporary hidden samples ${\mathbf{X}^\prime}$ and generated samples $\hat{\mathbf{X}}$ are utilized to perform the category loss $\mathcal{L}_{class}$.}
  \label{fig:top_arch}
\end{figure*}

Our framework is shown in Fig. \ref{fig:top_arch}, which mainly consists of three parts: the CNN feature extractor (FE), the hard sample generation (HSG) model, and the embedding learning loss functions. Formally, we denote the feature extraction CNN by $F$, which maps the input images $\mathbf{I}$ into embedding samples $\mathbf{X}$. We then use the proposed HSG model to further produce the hard samples $\hat{\mathbf{X}}$ based on the original samples $\mathbf{X}$. Note that the produced temporary hidden samples $\mathbf{X}^\prime$ in HSG are also applied in the loss computing process. The input images $\mathbf{I}$, the original embedding samples $\mathbf{X}$, the hidden samples $\mathbf{X}^\prime$ and the generated hard embedding samples $\hat{\mathbf{X}}$ share the same ground truth categories $\mathbf{L} = [l_1 , \cdots , l_n]$, where $l_i \in {1, \cdots , C}$. The goal is to train the deep embedding network $F$ with parameters $\theta_m$. 
The general metric learning approaches train the parameters $\theta_m$ through optimizing the following objective function:
\begin{equation}
    \theta_m^*=\mathop{\arg\min}_{\theta_m}  J(\theta_m; \mathbf{X}, F)
    \label{eq:F-train}
\end{equation}
We enhance the training procedure by utilizing the synthetic hard samples. Our final deep embedding learning metric is denoted as: 
\begin{equation}
    \theta_m^*=\mathop{\arg\min}_{\theta_m}  J(\theta_m; \mathbf{X},\mathbf{X}^\prime, \hat{\mathbf{X}},\mathbf{L},F)
    \label{eq:final-function}
\end{equation}
We rewrite (\ref{eq:final-function}) as:
\begin{align}
J&=
 \mathcal{L}_{F}=  w_o\mathcal{L}_{org} + w_l\mathcal{L}_{class} + w_h\mathcal{L}_{hard}
\label{eq:final-function1}
\end{align}
where $\mathcal{L}_{org}$, $\mathcal{L}_{class}$ and $\mathcal{L}_{hard}$ are loss terms corresponding to the original deep distance metric loss for $\mathbf{X}$, the category loss based on $\mathbf{L}$ and the deep distance metric loss for the generated hard sample $\hat{\mathbf{X}}$; $w_o$, $w_l$ and $w_h$ are three weighting factors corresponding to the three terms. Although the embedding model $F$ should be optimized on the whole dataset, we discuss the loss computing in a triplet input images $\mathbf{I}=\{I_a,I_p,I_n\}$ for convenience. In the following, we will detail the objective function (\ref{eq:final-function1}). 

\subsection{Objective Function}
For the input images $\mathbf{I}=\{I_a,I_p,I_n\}$, the corresponding triplet embedding samples $\mathbf{X} = \{\mathbf{a}, \mathbf{p}, \mathbf{n}\}$ will be produced by CNN feature extractor $F$. The hard triplet embedding samples $\hat{\mathbf{X}} = \{\hat{\mathbf{a}}, \hat{\mathbf{p}}, \hat{\mathbf{n}}\}$ are then further generated by passing $\mathbf{X}$ into HSG model. Besides, the temporary hidden embedding samples ${\mathbf{X}^\prime} = \{{\mathbf{a}^\prime}, {\mathbf{p}^\prime}\}$ are also obtained in HSG. The deep embedding network $F$ and the HSG model are simultaneously trained in an end-to-end manner. However, in the early stages of training, the deep model focuses on memorizing the simple training data \cite{zhang2016understanding}, and thus we can decrease the weight for the generated hard samples. Besides, in the early stage, the embedding space does not have an accurate semantic structure \cite{zheng2019hardness}, and so the quality of the generated hard embedding samples is very low and they can not provide valid information. Thus the use of the hard generated samples will ruin the learning and damage the embedding space structure. To solve this problem, we designed a weighting parameter adaptively adjusting method by using the training loss $\mathcal{L}_{G}$ of the HSG model. The embedding training objective function is represented as:
\begin{align}
 \mathcal{L}_{F}&=  w_o\mathcal{L}_{org} + w_l\mathcal{L}_{class} + w_h\mathcal{L}_{hard} \notag \\
 &=  e^{-\frac{\beta}{\mathcal{L}_{G}}}\mathcal{L}_{org} +\phi\mathcal{L}_{class} +(1-e^{-\frac{\beta}{\mathcal{L}_{G}}})\mathcal{L}_{hard} \notag \\
 &=  e^{-\frac{\beta}{\mathcal{L}_{G}}}\mathcal{L}_{t}(\mathbf{X}) + \phi\mathcal{L}_{sm}(\mathbf{X},\mathbf{X^{\prime}},\hat{\mathbf{X}},\mathbf{L}) + (1-e^{-\frac{\beta}{\mathcal{L}_{G}}})\mathcal{L}_{t}(\hat{\mathbf{X}})
\label{eq:final-function2}
\end{align}
where $\beta$ and $\phi$ are pre-defined parameters; $\mathcal{L}_{t}$ and $\mathcal{L}_{sm}$ are the triplet metric loss and the softmax loss, respectively; $e^{-\frac{\beta}{\mathcal{L}_{G}}}$, $\phi$, and $(1-e^{-\frac{\beta}{\mathcal{L}_{G}}})$ are three balancing parameters for the three terms: the original metric loss $\mathcal{L}_{org}$, the category loss $\mathcal{L}_{label}$ and the generated hard metric loss $\mathcal{L}_{hard}$. In the early stage of training, ${\mathcal{L}_{G}}$ is very large and thus the weight $e^{-\frac{\beta}{\mathcal{L}_{G}}}$ for $\mathcal{L}_{org}$ is larger than the weight $(1-e^{-\frac{\beta}{\mathcal{L}_{G}}})$ for $\mathcal{L}_{hard}$. As the training proceeds, ${\mathcal{L}_{G}}$ is decreased and thus high quality hard samples will be produced. The hard synthetics can provide information for training, and higher weight $(1-e^{-\frac{\beta}{\mathcal{L}_{G}}})$ is set to highlight the generated hard samples. Note that ${\mathcal{L}_{G}}$ is set as ${\mathcal{L}_{G_2}}$ in the HSG model, which will be discussed in Section \ref{sec:hsg}.
In (\ref{eq:final-function2}), $\mathcal{L}_{t}(\mathbf{X})$ and $\mathcal{L}_{t}(\hat{\mathbf{X}})$ are written as (\ref{eq:orix}) and (\ref{eq:genx}).
\begin{align}
\mathcal{L}_{t}(\mathbf{X}) 
&=\left[ \norm{\mathbf{a}-\mathbf{p}}^2_2-\norm{\mathbf{a}-\mathbf{n}}^2_2+\tau  \right]_+
\label{eq:orix}
\end{align}
\begin{align}
\mathcal{L}_{t}(\hat{\mathbf{X}}) 
&=\left[ \norm{\hat{\mathbf{a}}-\hat{\mathbf{p}}}^2_2-\norm{\hat{\mathbf{a}}-\hat{\mathbf{n}}}^2_2+\tau  \right]_+
\label{eq:genx}
\end{align}
The category loss is performed based on the softmax loss,
\begin{align}
 \mathcal{L}_{sm}(\mathbf{X},\mathbf{X^{\prime}},\hat{\mathbf{X}},\mathbf{L})
 =  -{\sum_{\mathbf{x}}^{\{\mathbf{a},\mathbf{p},\mathbf{n}\}}}l(\mathbf{x})\text{log}{\hat{l}(\mathbf{x})}  
 \nonumber\\-{\sum_{\mathbf{x}}^{\{\mathbf{a}^\prime,\mathbf{p}^\prime\}}}l(\mathbf{x})\text{log}{\hat{l}(\mathbf{x})} 
 -{\sum_{\mathbf{x}}^{\{\hat{\mathbf{a}},\hat{\mathbf{p}},\hat{\mathbf{n}}\}}}l(\mathbf{x})\text{log}{\hat{l}(\mathbf{x})}
\label{eq:softmax-label}
\end{align}
where $l(\mathbf{x})$ is the one-hot encoding vector of the correct class for sample $\mathbf{x}$; $\hat{l}(\mathbf{x})$ denotes the predicted probability vector.

In (\ref{eq:final-function2}), the only unknown elements are the generated samples $\hat{\mathbf{X}}$ and ${\mathbf{X}^\prime}$. In the following section, we will detail the hard sample generation. 

\section{{Two-stage Hard Sample Generation}}
\label{sec:hsg}

\begin{figure*}[htbp]
  \centering
  \setlength{\abovecaptionskip}{0.01cm} 
  \setlength{\belowcaptionskip}{-0.5cm}
  \includegraphics[width=1.0\linewidth]{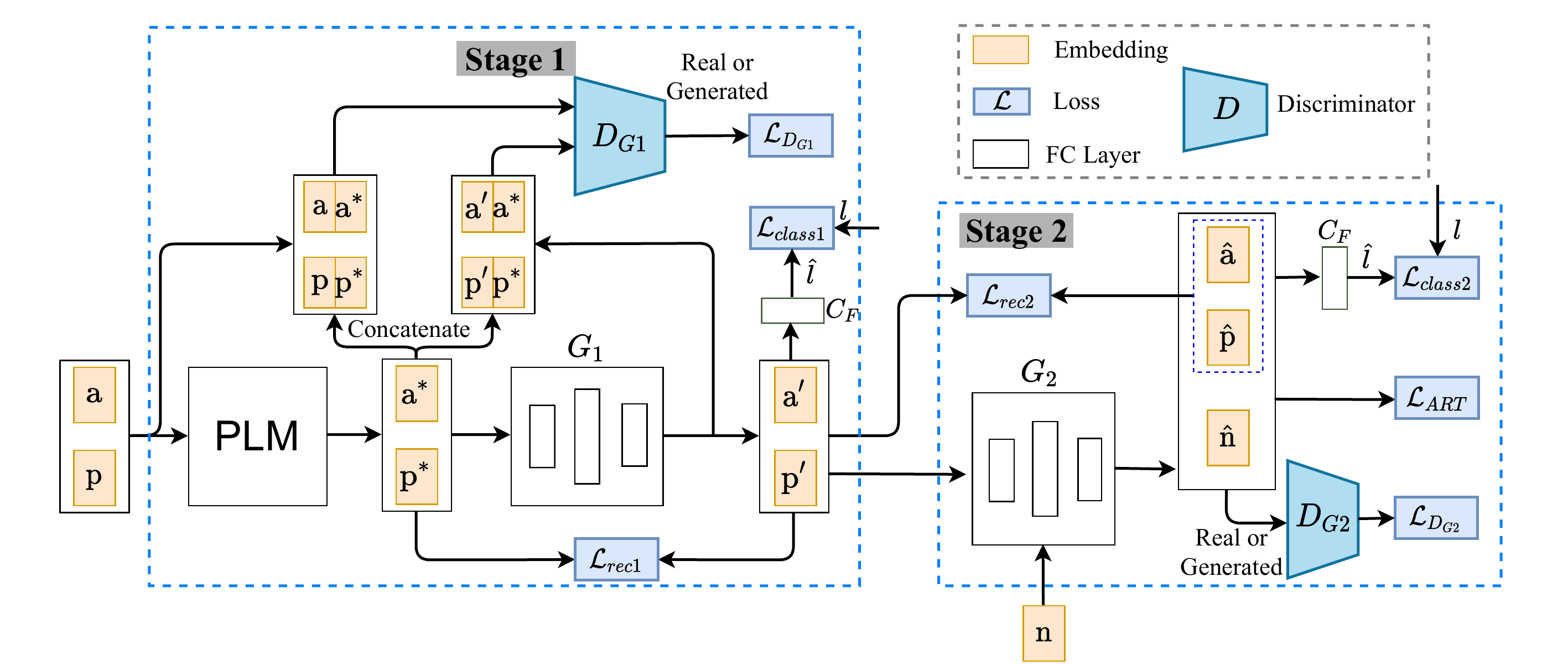}
  \caption{Hard sample generation architecture. The Stage 1 consists of one PLM module, one generator $G_1$ and one discriminator $D_{G1}$; the Stage 2 consists of one generator $G_2$ and one discriminator $D_{G2}$.}
  \label{fig:gen_arch}
\end{figure*}

\subsection{Architecture}

\begin{figure}[hbtp]
  \centering
  \includegraphics[width=3.46in]{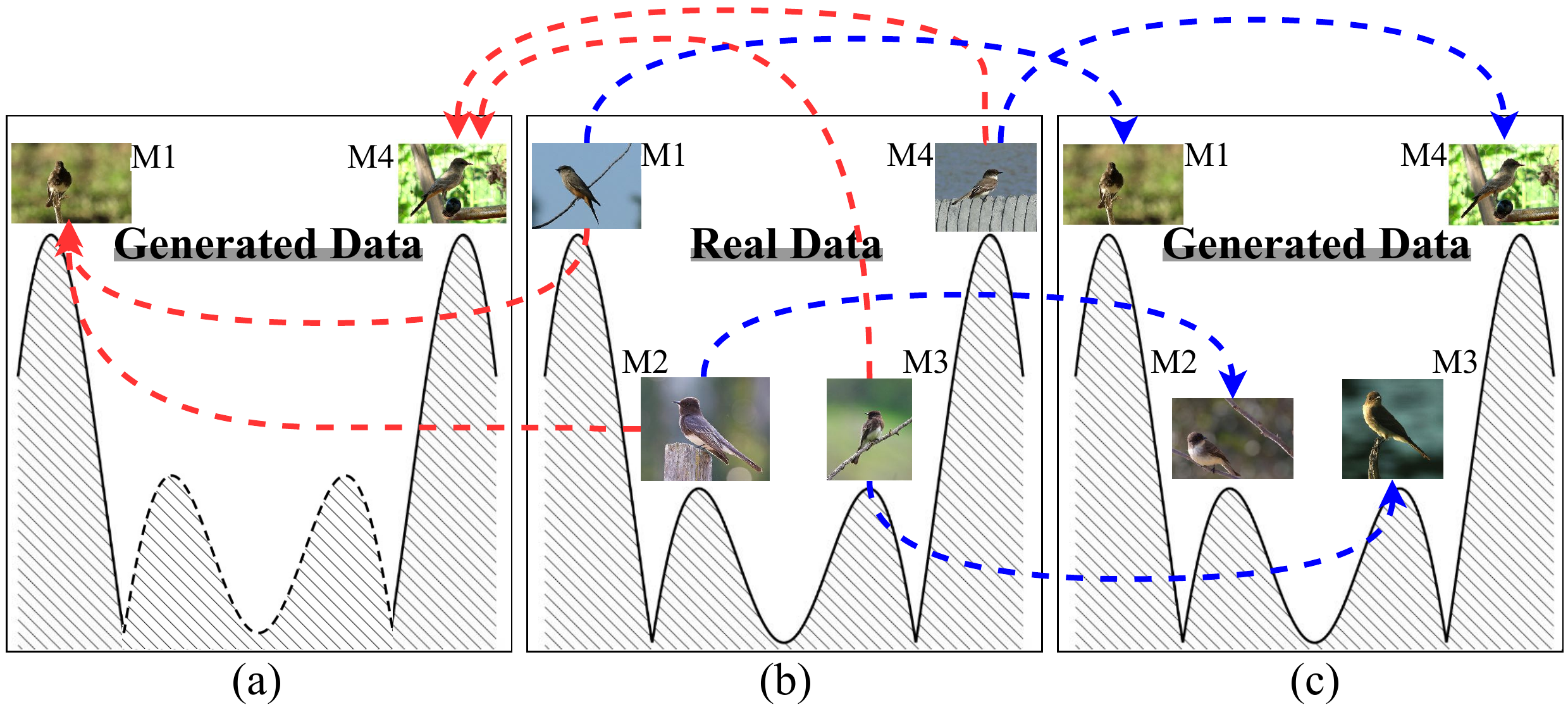}
  \caption{{Intuitive demonstration of label-preserving synthesis in work \cite{zheng2019hardness} and our proposed conditional synthesis. (a) label-preserving synthesis: generator may only produce samples from two modes to meet the requirement of the consistent label, resulting in mode collapse; (b) real data: the real data distribution contains samples of four modes with the same class label; (c) our conditional synthesis: we use a powerful conditional discriminator to alleviate the mode collapse issue by encouraging the one-to-one relationship between the original sample and the generated sample, and thus the diversity of the generated hard sample is improved.}}
  \label{fig:mode-collapse}
\end{figure}

The detailed two-stage hard sample generation architecture is denoted in Fig. \ref{fig:gen_arch}. In the first stage, the hard level of the input anchor-positive pair $\{\mathbf{a},\mathbf{p}\}$ will first be increased by directly manipulating the distances between them. To achieve this, the piecewise linear manipulation (PLM) is designed. Then the generated hard anchor-positive pair $\{\mathbf{a}^*,\mathbf{p}^*\}$ will be processed by generator $G_1$ and thus the hard anchor-positive pair $\{\mathbf{a}^{\prime},\mathbf{p}^{\prime}\}$ is obtained. However, just conducting label-preserving synthesis is not enough and this may result in mode collapse, as shown in Fig. \ref{fig:mode-collapse}(a). In this case, the generator only produces samples from part modes of the distribution and ignores the other modes. 

To alleviate the mode collapse issue and guarantee the diversity of the generated samples, we encourage the one-to-one relationship between the original sample and the generated sample by adding an extra conditional constraint to the generation process. The discriminator $D_{G1}$ needs to have the ability to identify the results of combining the intermediate embeddings with the original and generated embeddings respectively. In the second stage, the generated anchor-positive pair $\{\mathbf{a}^{\prime},\mathbf{p}^{\prime}\}$ and original negative sample $\mathbf{n}$ are passed into a generator $G_2$ to produce the final hard triplet $\{\hat{\mathbf{a}},\hat{\mathbf{p}},\hat{\mathbf{n}}\}$. To support the adversarial learning, the discriminator $D_{G2}$ is integrated into Stage 2. 

In our architecture, the PLM, the conditional synthesis (in Stage 1) and the hard triplet generation (in Stage 2) are three key parts that will be discussed in the following.

\subsection{Piecewise Linear Manipulation}
In order to make anchor $\mathbf{a}$ and positive $\mathbf{p}$ more difficult, we need to increase the distance between them. Here, we first linearly stretch the embeddings of anchor and positive samples, as shown in Stage 1 of Fig. \ref{fig:trip-gen}, thus generating a harder anchor and positive pair $\{\mathbf{a}^*, \mathbf{p}^*\}$, as depicted in Fig. \ref{fig:gen_arch}. We propose the piecewise linear manipulation (PLM) scheme as shown in (\ref{eq:pulling}),
\begin{equation}
\begin{split}
\mathbf{a}^*=\mathbf{a}+\lambda(\mathbf{a}-\mathbf{p}) \\
\mathbf{p}^*=\mathbf{p}+\lambda(\mathbf{p}-\mathbf{a}) 
    \label{eq:pulling}
\end{split}
\end{equation}
where $\lambda$ is a picecwise variable which controls the hard level of the generated anchor-positive pair $\{\mathbf{a}^*, \mathbf{p}^*\}$. The picecwise variable $\lambda$ plays the key role in our PLM scheme and we will discuss it in the following.

During the stretching process, if the value of $\lambda$ is too large, the anchor-positive samples may be stretched into different classes. Even if the distance is increased, the stretched samples will no longer be anchor-positive pair and thus become meaningless. In order to avoid this issue, we need to limit the scope and size of the $\lambda$. Thus in our PLM, we set the piecewise variable $\lambda$ as: 
\begin{equation}
	\lambda = \begin{cases}
	\frac{\alpha}{e^{d(\mathbf{a},\mathbf{p})-d_t}}, & {if} \ \ \ d(\mathbf{a},\mathbf{p}) \geq d_t  \\
  \alpha + \gamma(1 - \frac{d(\mathbf{a},\mathbf{p})}{d_t}), & if \ \ \ d(\mathbf{a},\mathbf{p}) < d_t 
		   \end{cases}
\label{eq:pulling-piece}
\end{equation}
where $d(\mathbf{a},\mathbf{p})$ denotes the distance between the anchor and positive samples; $d_t$ is an adaptive threshold. When $d(\mathbf{a},\mathbf{p})$ is already large enough to be greater than $d_t$, we use a exponent function ${\alpha}/{e^{d(\mathbf{a},\mathbf{p})-d_t}}$. At this situation, the maximum value of $\lambda$ is $\alpha$, and the minimum value is about 0. When the distance between them is less than $d_t$, we use a linear function $\alpha + \gamma(1 - \frac{d(\mathbf{a},\mathbf{p})}{d_t})$. At this situation, the maximum value of $\lambda$ is $\alpha+\gamma$ and the minimum value is $\alpha$.
In the calculation of $\lambda$, $d_t$ is related to the distance distribution of samples in different datasets, so it is difficult to manually adjust $d_t$ in various datasets. In order to automatically select the best hyperparameters $d_t$ in each dataset, we proposed an adaptive method by choosing $d_t$ as: 
\begin{equation}
	d_t=\frac{1}{M}\sum_{i=1}^M{d_{epoch-1}(\mathbf{a}_i,\mathbf{p}_i)}
\label{eq:d0-set}
\end{equation}
where $M$ is the number of anchor-positive pairs, and $d_t$ represents the average distance of anchor-positive pairs in the previous epoch. 

After piecewise linearly manipulating the input anchor-positive pair $\{\mathbf{a},\mathbf{p}\}$, we obtain hard anchor-positive pair $\{\mathbf{a}^*,\mathbf{p}^*\}$. However, there is no guarantee that $\{\mathbf{a}^*,\mathbf{p}^*\}$ share the same label with $\{\mathbf{a},\mathbf{p}\}$. Next, we will conduct conditional synthesis based on the generated $\{\mathbf{a}^*,\mathbf{p}^*\}$ to produce valid hard anchor-positive pair $\{\mathbf{a}^\prime,\mathbf{p}^\prime\}$.

\subsection{Conditional Synthesis}
After piecewise linearly manipulating anchor-positive pairs, in order to ensure the generated pairs are still in the same category domain as the original samples, the recent method requires label consistency to avoid generating invalid embeddings \cite{duan2018deep}. However, the simple constraint may result in a mode collapse problem, which means the generated samples suffering from insufficient diversity, as shown in Fig \ref{fig:mode-collapse}. 
Thus except for the label-preserving requirement, we try to enhance the relationship between the original sample and the generated sample by adding an extra restriction. Specifically, we require that the generated samples $\{\mathbf{a}^{\prime},\mathbf{p}^{\prime}\}$ based on $\{\mathbf{a}^*,\mathbf{p}^*\}$ have the same distribution as the original samples $\{\mathbf{a},\mathbf{p}\}$. Unlike an unconditional generative adversarial network, both the generator $G_1$ and discriminator $D_{G1}$ consider the information of the 
linear-manipulated anchor-positive pair $\{\mathbf{a}^*,\mathbf{p}^*\}$. To summarize, we encourage the one-to-one relationship between the original sample and the generated sample for our conditional synthesis, and thus the diversity is ensured. 

When $\{\mathbf{a}^*,\mathbf{p}^*\}$ passed $G_1$, they are mapped to $\{\mathbf{a}^\prime,\mathbf{p}^\prime\}$. On the one hand, we synthesize anchor-positive pairs that are further apart based on $\{\mathbf{a}^*,\mathbf{p}^*\}$; On the other hand, no matter where the input $x^{*}$ of $G_1$ lies in its category space, the output of $G_1$ is located in the same category space as the original one by the constraint of label consistency. Our conditional synthesis is achieved by training $G_1$ with the following loss function:
\begin{equation}
    \begin{aligned}
        \mathcal{L}_{G1} &= \eta(\mathcal{L}_{class1}+\mathcal{L}_{D_{G1},adv})+{(1-2\eta)}\mathcal{L}_{rec1}\\
    \end{aligned}
    \label{eq:HAPG-loss}
\end{equation}
where $\eta$ is the balance factor between softmax and adversarial loss; the first term makes sure the generated embeddings have the same labels with the original ones; the second term is adversarial loss with discriminator $D_{G1}$ which aims to let the generated embeddings look real, not the fake ones; the third term guarantees the generated embeddings can maintain the same hard level as the input embedings. 
In (\ref{eq:HAPG-loss}), the first term is denoted as:

\begin{equation}
    \begin{aligned}
\mathcal{L}_{class1}&={\sum_{\mathbf{x}^{\prime}}}\mathcal{L}_{sm}\!(C_{F}(\mathbf{x}^{\prime}),\!l_{\mathbf{x}^{\prime}})
    \end{aligned}
    \label{eq:dg1label-loss}
\end{equation}
where $\mathbf{x}^\prime$ denotes embeddings from $\{\mathbf{a}^{\prime},\mathbf{p}^{\prime}\}$ and $l_{\mathbf{x}^{\prime}}$ is the ground truth label of $\mathbf{x}^{\prime}$; the $\mathcal{L}_{sm}$ denotes the {softmax loss}; $C_F$ is a shared fully connected layer which is used to classify the generated pair. In (\ref{eq:HAPG-loss}), the second term is denoted as:

\begin{equation}
    \begin{aligned}
\mathcal{L}_{D_{G1},adv}&={\sum_{\mathbf{{x}^{\prime}_{c}}=[\mathbf{x}^{\prime}, \mathbf{x}^*]}}\mathcal{L}_{sm}(D_{G1}(\mathbf{x}^{\prime}_{c}),1)
    \end{aligned}
    \label{eq:DG1adv-loss}
\end{equation}
where $\mathbf{x}$ denotes embeddings from $\{\mathbf{a},\mathbf{p}\}$, $\mathbf{x}^*$ denotes embeddings from $\{\mathbf{a}^*,\mathbf{p}^*\}$ and "1" represents the real samples. The discriminator $D_{G1}$ takes $\mathbf{x}^{\prime}_{c}$ the result of concatenating $\mathbf{x}^{\prime}$ and $\mathbf{x}^*$ along the last dimension as the input. This adversarial loss is used to fool the discriminator $D_{G1}$ by generating real-like samples; the 
third term in (\ref{eq:HAPG-loss}) is reconstruction loss $\mathcal{L}_{rec1}$ between $\{\mathbf{a}^{*},\mathbf{p}^{*}\}$ and $\{\mathbf{a}^{\prime},\mathbf{p}^{\prime}\}$, which is calculated as:
 \begin{equation}
     \begin{aligned}
	\mathcal{L}_{rec1}={d(\mathbf{a^*},\mathbf{a}^{\prime})+d(\mathbf{p^*},\mathbf{p}^{\prime})} = \norm{\mathbf{a^*}-\mathbf{{a}^{\prime}}}^2_2 + \norm{\mathbf{p^*}-\mathbf{{p}^{\prime}}}^2_2.
    \end{aligned}
\end{equation}
With the reconstruction loss $\mathcal{L}_{rec1}$, the relative hardness of anchor-positive synthetic pairs in triplets will be guaranteed. In this way, our generative loss $\mathcal{L}_{G1}$ enables generated anchor-positive pairs to have a greater separation distance while remaining within their original category space.

The discriminator $D_{G1}$ is a two-category classifier for identifying whether a given embedding is a real one or a generated one. The training loss function for $D_{G1}$ is formulated as:
\begin{equation}
    \begin{aligned}
    \mathcal{L}_{D_{G1}} = \frac{1}{2}\{{\sum_{\mathbf{x}_{c}=[\mathbf{x}, \mathbf{x}^*]}}\mathcal{L}_{sm}(D_{G1}(\mathbf{x}_{c}), 1)
    + {\sum_{\mathbf{{x}^{\prime}_{c}}=[\mathbf{x}^{\prime}, \mathbf{x}^*]}}\mathcal{L}_{sm}(D_{G1}(\mathbf{{x}^{\prime}_{c}}), 0)\}
    \end{aligned}
    \label{eq:HAPDloss}
\end{equation}
where $\mathbf{x}$, $\mathbf{x}^*$ and $\mathbf{x}^\prime$ denote embeddings from $\{\mathbf{a},\mathbf{p}\}$, $\{\mathbf{a}^*,\mathbf{p}^*\}$ and $\{\mathbf{a}^\prime,\mathbf{p}^\prime\}$, respectively. $\mathbf{x}_{c}$ and $\mathbf{{x}^{\prime}_{c}}$, the input embedding of the discriminator $D_{G1}$, are concatenated of $\mathbf{x}^*$ and corresponding embedding in the last dimension of the same labeled instance. "1" and "0" denote the real sample and the fake sample respectively.

After the adversarial training of $G_1$, our hard anchor-positive pairs have met the requirements of label consistency. At the same time, conditional discriminator $D_{G1}$ ensures that the generated pairs do not have mode collapse with random generation. Next, we will use the hard anchor-positive pair $\{\mathbf{a}^\prime,\mathbf{p}^\prime\}$ to generate the hard negative, and produce the final hard samples $\{\hat{\mathbf{a}},\hat{\mathbf{p}},\hat{\mathbf{n}}\}$.

\subsection{Hard Triplet Synthesis}
We propose a new hard triplet generator $G_2$ to generate hard samples. Generator $G_2$ is able to map its input $\{\mathbf{a}^\prime,\mathbf{p}^\prime\}$ and $\mathbf{n}$ to $\{\hat{\mathbf{a}},\hat{\mathbf{p}},\hat{\mathbf{n}}\}$. While generating a negative sample, the hard negative sample is not only related to the information of negative sample itself, but also to the anchor-positive pair in the triplet. Therefore the inputs of $G_2$ are negative sample $\mathbf{n}$ and hard anchor-positive pair $\{\mathbf{a}^{\prime}, \mathbf{p}^{\prime}\}$ generated in Stage 1. For our hard triplet synthesis, we propose the loss function to train $G_2$ as follows:
\begin{equation} 
    \begin{aligned}
        \mathcal{L}_{G2} = \mu  \mathcal{L}_{ART} \!+\! (1\!\!-\!2\eta\!-\!\!\mu)  \mathcal{L}_{rec2}\!+\!\eta  ( \mathcal{L}_{class2} \!+\mathcal{L}_{DG2,adv} \!\!)
    \end{aligned}
    \label{eq:HTG-loss1}
\end{equation}
where $\eta$ and $\mu$ are the balancing factors. The overall objective loss function of $G_2$ is composed of four parts: the adaptive reverse triplet loss (ART-loss) $\mathcal{L}_{ART}$, reconstruction loss $\mathcal{L}_{rec2}$, softmax loss $\!\mathcal{L}_{class2}\!$ and adversarial loss $\mathcal{L}_{DG2,adv} \!$. The ART-loss is used to generate the hard negative; the reconstruction loss is applied to keep hard anchor-positive pair from being affected by reverse triplet loss; the softmax loss is used to ensure label consistency for the generated samples; adversarial loss with discriminator $D_{G2}$ is performed to generate samples like the real ones. These four terms are detailed in the following.


The proposed ART-loss $\mathcal{L}_{ART}$ is defined as follows:
\begin{equation}
\begin{aligned}
    \mathcal{L}_{ART}=\mathcal{L}_\text{r}(\mathbf{a}^{\prime},\mathbf{p}^{\prime},\mathbf{n})
    = \left[ \left\|G_2(\mathbf{a^{\prime}})\!\!-\!\!G_2(\mathbf{n})\right\|^2_2\!\!-\left\|G_2(\mathbf{a}^{\prime})\!\!-\!\!G_2(\mathbf{p}^{\prime})\right\|^2_2+\tau_r  \right]_+ \\
\end{aligned}
\label{eq:reverse-tri}
\end{equation}
The proposed ART-loss is obtained from the triplet loss by reversing the sign in front of the first two terms in (\ref{eq:trip}). The purpose of ART-loss is to make the distance between $\mathbf{a}^{\prime}$ and $\mathbf{n}$ as small as possible until it is less than the distance between $\mathbf{a}^{\prime}$ and $\mathbf{p}^{\prime}$ with a parameter $\tau_r$. When $\mathbf{n}$ satisfies this condition, it will be a hard negative sample. In this paper, we propose to use an adaptive parameter $\tau_r$ which increases with the training of the network. We impose more stringent restrictions on reverse triplet loss with larger $\tau_r$ when the performance of $G_2$ is getting better. Thus, we can gradually generate harder and harder negative samples. We set $\tau_r$ as a parameter that changes with $\mathcal{L}_{G2}$ as:
\begin{equation}
    \tau_r = \nu(1 -\exp^{-\frac{\beta}{\mathcal{L}_{G2}}})
    \label{eq:tau}
\end{equation}
where $\nu$ and $\beta$ are constant parameters. As $G_2$ trains better, the $\mathcal{L}_{G2}$ will get smaller, and $\tau_r$ will increase adaptively to enhance the difficulty of hard negative.

The reconstruction loss $\mathcal{L}_{rec2}$ aims to keep the synthetic anchor-positive pair with a consistent hard-level  by decreasing the distances between $\{\mathbf{a}^\prime,\mathbf{p}^\prime\}$ and $\{\hat{\mathbf{a}},\hat{\mathbf{p}}\}$, which is denoted as: 
 \begin{equation}
	\mathcal{L}_{rec2}={d(\mathbf{a}^{\prime},\hat{\mathbf{a}})+d(\mathbf{p}^{\prime},\hat{\mathbf{p}})} \\
	= \norm{\mathbf{a}^{\prime}-\hat{\mathbf{a}}}^2_2 + \norm{\mathbf{p}^{\prime}-\hat{\mathbf{p}}}^2_2
\label{eq:d1-set}
\end{equation}
The softmax loss $\!\mathcal{L}_{class2}\!$ makes the generated embeddings have the same labels with the original samples, which is calculated as:
\begin{equation}
    \begin{aligned}
        \mathcal{L}_{class2} &= {\sum_{\hat{\mathbf{x}}}} \mathcal{L}_{sm}\!(C_{F}(\hat{\mathbf{x}}), \!l_{\hat{\mathbf{x}}})
    \end{aligned}
    \label{eq:label2-loss}
\end{equation}
where $\hat{\mathbf{x}}$ denotes embeddings from $\{\hat{\mathbf{a}},\hat{\mathbf{p}},\hat{\mathbf{n}}\}$, and $l_{\hat{\mathbf{x}}}$ denotes the ground truth labels of $\hat{\mathbf{x}}$.

The adversarial loss with discriminator $D_{G2}$ is
\begin{equation}
    \begin{aligned}
        \mathcal{L}_{D_{G2},adv}
          &= {\sum_{\hat{\mathbf{x}}}} \mathcal{L}_{sm}\!(D_{G2}(\hat{\mathbf{x}}), \!l_{\hat{\mathbf{x}}})
    \end{aligned}
    \label{eq:dg2adv-loss}
\end{equation}
Different with the previous adversarial loss, such as (\ref{eq:DG1adv-loss}), we use the ground truth label $l_{\hat{\mathbf{x}}}$ to represent the ``real" samples. Thus the adversarial loss (\ref{eq:dg2adv-loss}) is used to fool the discriminator by classifying these generated embeddings as the real samples with ground truth $l_{\hat{\mathbf{x}}}$. Actually, our adopted discriminator $D_{G2}$ is a $(C+1)$-class classifier where $C$ denotes the number of ground truth categories for the real samples and the left one class represents the generated samples. When an embedding is given, $D_{G2}$ can discriminate it into the $C + 1$ categories \cite{mirza2014conditional} and thus the real-like sample generation and label consistency are achieved at the same time. The training loss for $D_{G2}$ is denoted as:  
\begin{equation}
    \mathcal{L}_{D_{G2}} = \frac{1}{C+1}\{{\sum_{\mathbf{x}^{\prime}}}\mathcal{L}_{sm}(D_{G2}(\mathbf{x}^{\prime}), l_{{\mathbf{x}}^{\prime}}) + {\sum_{\hat{\mathbf{x}}}}\mathcal{L}_{sm}(D_{G2}(\hat{\mathbf{x}}), 0)\}
    \label{eq:D2loss}
\end{equation}
where $\mathbf{x}^\prime$ denotes embeddings from $\{\mathbf{a}^\prime,\mathbf{p}^\prime\}$ or $\mathbf{n}$, $l_{{\mathbf{x}}^\prime}$ denotes the ground truth labels of ${\mathbf{x}}^\prime$, and $\hat{\mathbf{x}}$ denotes embeddings from $\{\hat{\mathbf{a}},\hat{\mathbf{p}},\hat{\mathbf{n}}\}$.

\section{Experiment}
\label{sec:experiment}
{In this section, extensive experiments are performed to demonstrate the superior performance of our proposed hard sample generation. Besides, an ablation study is further developed to show the contributions of each component of our proposed algorithm.}

\subsection{Dataset}
\label{sec:data}

We evaluated our method under the three datasets below. In order to better compare with other methods, we split the datasets according to \cite{duan2018deep}\cite{zheng2019hardness}\cite{oh2016deep}\cite{oh2017deep}, and the training set and test set contain image classes with no intersection.
\begin{itemize}
\item The CUB-200-2011 dataset \cite{wah2011caltech} consists of 200 bird species with 11,788 images. We use the first 100 species (5,864 images) for training and the rest 100 species (5,924 images) for testing.
\item The Cars196 dataset \cite{krause20133d} consists of 196 car makes and models with 16,185 images. We use the first 98 models (8,054 images) for training and the rest 100 models (8,131 images) for testing.
\item The Stanford Online Products dataset (SOP) \cite{oh2016deep} consists of 22,634 online products from eBay.com with 120,053 images. We use the first 11,318 products (59,551 images) for training and the rest 11,316 products (60,502 images) for testing.
\end{itemize}

\subsection{Implementation}
\label{sec:impl}

\textbf{Configuration.} The proposed method is implemented with Python3.7 and PyTorch 1.6.0, using the
GoogleNet \cite{szegedy2015going} and ResNet50 \cite{he2016deep} architectures, pre-trained on ImageNet ILSVRC dataset \cite{russakovsky2015imagenet}. All experiments are performed on individual NVIDIA Tesla T4 GPUs. We trained networks with standard back-propagation, which is performed by Adam optimizer with 4e-4 weight decay. We set the initial learning rate 1e-5 for the feature extractor, 1e-3 for the classifier, 1e-4 for the discriminators, and 1e-3 for the generators. Each generator is composed of two fully connected layers with dimensions 128 and 512. Note that all output embedding vectors of the feature extractor and generators are under normalization. Each training is run over 150 epochs for CUB200-2011/CARS196 and 100 epochs for Stanford Online Products. For a fair comparison with previous methods on deep metric learning, we used GoogLeNet \cite{szegedy2015going} architecture as the CNN feature extractor with output embedding size of 512 for all the three datasets similar to \cite{zheng2019hardness} and \cite{duan2018deep}. While training, we set the batch size to 128, and all the training images are resized to $227 \times 227$. During the training process, we will update all discriminators, generators, and the embedding extraction network according to Algorithm \ref{alg:algorithm1}. We set triplet margin $\tau= 0.2$ and set the the maximum value of  reverse triple margin $\nu=0.2$ as well. The other related parameters of this work are summarized in Table \ref{tab:parameter-test}.

\begin{table}[htp]
	\centering
	\begin{tabular}{l ccc ccc cc}
        \toprule
		\textbf{}& \textbf{$\alpha$} & \textbf{$\gamma$} & \textbf{$\eta$} & \textbf{$\beta$}  & \textbf{$\mu$} & \textbf{$\phi$} 
		\\ \midrule
		\textbf{Cub200} & 0.2 & 0.8 & 0.3 & 0.5 & 0.3 & 0.5  \\
		\textbf{Cars196}  & 0.2 & 0.8 & 0.3 & 0.5 & 0.3 & 0.25  \\
		\textbf{SOP}     & 0.2 & 0.8 & 0.3 & 0.5 & 0.3 & 0.2  \\
		\bottomrule
	\end{tabular}
	\caption{The related parameter settings for different datasets.}
	\label{tab:parameter-test}
\end{table}

\textbf{Training Flow.} We alternately train $F$ and the other models, and mini-batch SGD is applied in the training process. Specifically, we first initialize all the models and then pre-train model $F$. After that, we will iteratively train all the models for $epochs$ times. In each epoch, these models are updated for $batches$ times. For a specific batch, embeddings $\mathbf{X}$ are first produced by $F$, and then $\mathbf{X^*}$ and $\mathbf{X^{\prime}}$ can be obtained by using PLM and $G_1$. Then model ${D_{G_1}}$ can be updated with (\ref{eq:HAPDloss}). After that, the generator ${{G_1}}$ are updated with (\ref{eq:HAPG-loss}) accordingly. However, for the training of $G_2$ and $D_{G_2}$, we will use the samples produced by the latest generative networks. For this reason, we will re-extract embeddings $\mathbf{X^{\prime}}$ after the model updating of Stage 1 for the training of $G_2$ and $D_{G_2}$ in Stage 2. Then we update them with (\ref{eq:D2loss}) and (\ref{eq:HTG-loss1}) accordingly. Finally, we update model $F$ with the re-extracted $\hat{\mathbf{X}}$, $\mathbf{X^{\prime}}$ and the original $\mathbf{X}$ by using (\ref{eq:final-function2}). With the above training flow, the final embedding model $F$ and the other models related to hard sample generation are obtained at the same time. When our network training flow is over, there is no additional computational effort involved in the final feature extraction for image retrieval. We summarize the whole embedding training flow as Algorithm \ref{alg:algorithm1}. 

\begin{algorithm}[bt]
\caption{embedding training flow}
\label{alg:algorithm1}
\textbf{Input}: model $F$, $G_1$, $D_{G_1}$, $G_2$, $D_{G_2}$, images $\mathbf{I}$, labels $\mathbf{L}$,  \\
\textbf{Parameter}: $epochs$,    \\
\textbf{Output}: trained $F$ 
\begin{algorithmic}[1] 
\STATE \textbf{init} model $F$, $G_1$, $D_{G_1}$, $G_2$, $D_{G_2}$
\STATE pre-train model $F$
\FOR{$i=1$ \TO $epochs$} 
\FOR{$j=1$ \TO $batches$} 
\STATE extract embeddings $\mathbf{X}=F(\mathbf{I}_{\text{batch}_j})$, $\mathbf{X^*}$ and $\mathbf{X^{\prime}}$,
\STATE update ${{G_1}}$ and ${D_{G_1}}$ with (\ref{eq:HAPG-loss}) and (\ref{eq:HAPDloss}) respectively,
\STATE re-extract embeddings $\mathbf{X}^{\prime}$,
\STATE extract embeddings $\hat{\mathbf{X}}$,
\STATE update ${{G_2}}$ and ${D_{G_2}}$ with (\ref{eq:D2loss}) and (\ref{eq:HTG-loss1}) respectively,
\STATE re-extract embeddings $\hat{\mathbf{X}}$,
\STATE update model ${{F}}$ with (\ref{eq:final-function2}),
\ENDFOR
\ENDFOR
\STATE \textbf{return} model $F$
\end{algorithmic}
\end{algorithm}

\subsection{Evaluation}
We designed experiments to prove the effectiveness of our hard-sample generation.

\textbf{Evaluation Criteria.}
We report the performance for both retrieval and clustering tasks. The mAP is widely used in state-of-the-art image retrieval works and it reveals the position-sensitive ranking precision. The definition of mAP is denoted as:
\begin{equation}\label{map}
\text{mAP} = \frac{1}{N}\sum_{i=1}^{N}\frac{\sum_{r=1}^{M^i_\text{relevant}}P(r)}{M^i_\text{relevant}}   
\end{equation}
where $i$ is the $i_{th}$ query index and $N$ is the number of total queries \cite{zhu2019feature}. $M^i_\text{relevant}$ is the number of relevant images corresponding to the $i_{th}$ query, ${r}$ is the rank and $P(r)$ is the precision at the cut-off rank of $r$. The employed Recall@K metric is determined by the existence of at least one correct retrieved sample in the K nearest neighbors \cite{zheng2019hardness}.

For the clustering task, we employed NMI and $F_1$ as performance metrics. The normalized mutual information (NMI) \cite{oh2016deep} is defined by the ratio of the mutual information of clusters and ground truth labels and the arithmetic mean of their entropy: $\operatorname{NMI}(\Omega, \mathbb{C})=\frac{I(\Omega ; \mathbb{C})}{H(\Omega)+H(\mathbb{C})}$, where $\Omega=\left\{\omega_{1}, \ldots, \omega_{n}\right\}$ is a set of clusters and $\mathbb{C}=\left\{c_{1}, \ldots, c_{n}\right\}$ is the ground truth classes. Here $I(\cdot,\cdot)$ and $H(\cdot)$ denotes mutual information and entropy respectively. $F_1$ score is deined as the harmonic mean of precision and recall \cite{oh2016deep}.

\textbf{Qualitative Analysis.} We visualize our image retrieval performance in several selected cases and visually show the advantage of our algorithm. Besides, we depict the learned embeddings for different datasets and qualitatively verify the effectiveness of the proposed scheme.

\subsection{Objective Comparison}
\begin{table}[htb]
	\centering
	\begin{tabular}{lccccccc}
	\toprule
    Method  & R@1  & R@2  & R@4  & R@8  & NMI  & F1   & mAP  \\ \midrule
    Triplet\cite{zheng2019hardness} & 35.9 & 47.7 & 59.1 & 70.0 & 49.8 & 15.0 & 19.2 \\
    MAC     & 40.3 & 51.9 & 63.6 & 74.1 & -    & -    & 20.1 \\
    R-MAC   & 41.7 & 53.7 & 66.3 & 76.3 & -    & -    & 20.8 \\
    GeM     & 42.2 & 54.3 & 67.1 & 77.4 & -    & -    & 21.4 \\
    DAML\cite{duan2018deep}    & 37.6 & 49.3 & 61.3 & 74.4 & 51.3 & 17.6 & 21.7 \\
    HDML\cite{zheng2019hardness}    & 43.6 & 55.8 & 67.7 & 78.3 & 55.1 & 21.9 & 22.8 \\
    SS\cite{gu2020symmetrical}    & 51.4 & 63.0 & 74.4 & 84.1 & 59.6 & 26.2 & - \\
    OURS$^{128}$ &  \textcolor{blue}{55.3} &  \textcolor{blue}{66.6} &  \textcolor{blue}{77.3} &  \textcolor{blue}{86.0} &  \textcolor{blue}{62.6} &  \textcolor{blue}{31.5} &  \textcolor{blue}{26.8} \\
    OURS    & \textbf{57.0} & \textbf{68.8} & \textbf{79.2} & \textbf{86.7} & \textbf{64.1} & \textbf{33.0} & \textbf{27.9} \\    \bottomrule
	\end{tabular}
	\caption{Experimental results on the CUB-200-2011 dataset in comparison with other methods(\%).}
	\label{cub-result}
\end{table}

\begin{table}[htbp]
	\centering
	\begin{tabular}{lccccccc}
	\toprule
    Method  & R@1  & R@2  & R@4  & R@8  & NMI  & F1   & mAP  \\  \midrule
    Triplet\cite{zheng2019hardness} & 45.1 & 57.4 & 69.7 & 79.2 & 52.9 & 17.9 & 20.8 \\
    MAC     & 56.5 & 69.3 & 79.1 & 87.2 & -    & -    & 17.0 \\
    R-MAC   & 60.0 & 71.8 & 81.1 & 88.2 & -    & -    & 20.5 \\
    GeM     & 60.4 & 71.9 & 81.2 & 88.1 & -    & -    & 21.5 \\
    DAML\cite{duan2018deep}    & 60.6 & 72.5 & 82.5 & 89.9 & 56.5 & 22.9 & 20.7 \\
    HDML\cite{zheng2019hardness}    & 61.0 & 72.6 & 80.7 & 88.5 & 59.4 & 27.2 & 22.5 \\
    SS\cite{gu2020symmetrical}    & 69.7 & 78.7 & 86.1 & 91.4 & 62.4 & 31.8 & -    \\
    OURS$^{128}$  & \textcolor{blue}{79.3} & \textcolor{blue}{86.9} & \textcolor{blue}{92.2} & \textcolor{blue}{95.7} & \textcolor{blue}{65.5} & \textcolor{blue}{35.4} & \textcolor{blue}{29.6} \\
    OURS    & \textbf{82.1} & \textbf{88.7} & \textbf{93.1} & \textbf{96.0} & \textbf{66.3} & \textbf{35.1} & \textbf{31.1} \\
		\bottomrule
	\end{tabular}
	\caption{Experimental results on the Cars196 dataset in comparison with other methods(\%).}
	\label{Cars196-result}
\end{table}

\begin{table}[htb]
	\centering
    \begin{tabular}{lcccccc}
		\toprule
    Method  & R@1  & R@10 & R@100 & NMI  & F1   & mAP  \\    \midrule
    Triplet\cite{zheng2019hardness} & 53.9 & 72.1 & 85.7  & 20.2 & 15.0 & 29.7 \\
    MAC     & 54.2 & 72.5 & 85.9  & -    & -    & 30.5 \\
    R-MAC   & 55.4 & 73.1 & 86.1  & -    & -    & 31.3 \\
    GeM     & 56.7 & 74.3 & 86.7  & -    & -    & 32.8 \\
    DAML\cite{duan2018deep}    & 58.1 & 75.0 & 88.0  & 87.1 & 22.3 & 34.9 \\
    HDML\cite{zheng2019hardness}    & 58.5 & 75.5 & 88.3  & 87.2 & 22.5 & \textbf{35.7} \\
    SS\cite{gu2020symmetrical}    & 65.7 & \textcolor{blue}{81.4} & \textbf{91.7}  & \textbf{88.9} & \textbf{30.6} & - \\
    OURS$^{128}$ & \textcolor{blue}{66.1} & 80.8 & 90.8  & 87.6 & 25.4 & 33.0  \\
    OURS    & \textbf{67.7} & \textbf{82.2} & \textcolor{blue}{91.6} & \textcolor{blue}{87.7} & \textcolor{blue}{26.1} & \textcolor{blue}{34.4} \\
		\bottomrule
	\end{tabular}
	\caption{Experimental results on the SOP dataset in comparison with other methods(\%).}
	\label{online-result}
\end{table}

\begin{figure}[htb]
  \centering
  \includegraphics[width=3.40in]{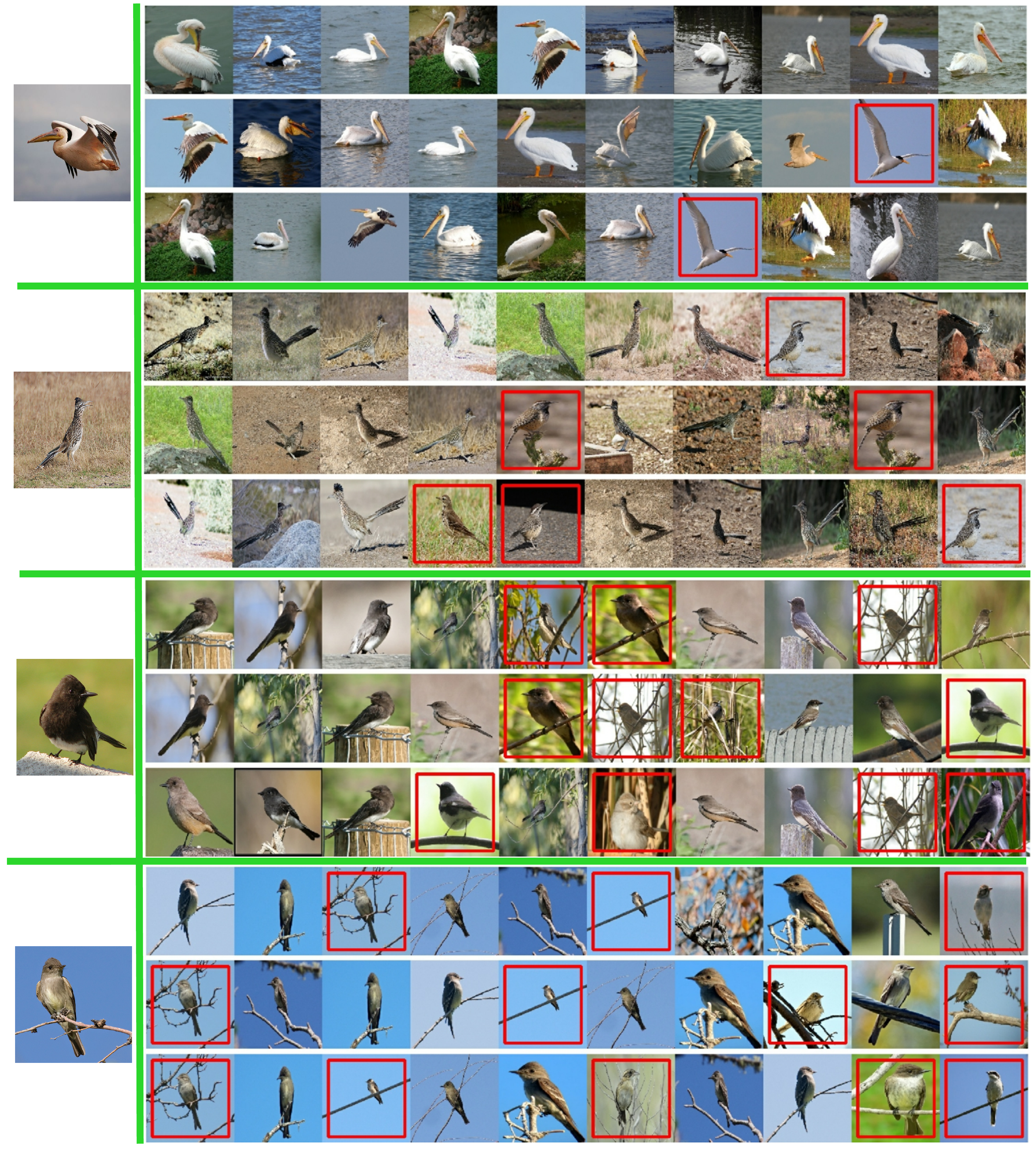}
  \caption{{Visualized case study of image retrieval. The first image in each group (including 3 rows) is the query, and the 3 rows of retrieval results are our method, work \cite{zheng2019hardness}, and the basic triplet embedding based retrieval. The red rectangle indicates an incorrect retrieval image.}}
  \label{fig:retrieval-vis}
\end{figure}

{We conduct full comparisons between our method and the other image retrieval algorithms. The involved strategies include deep pooling features and the recent deep embeddings with hard generation. For a fair comparison, all the involved algorithms are implemented with triplet loss architecture.

\textbf{Typical Image Retrieval Methods.} To comprehensively demonstrate the superiority of our proposed method over existing methods in the retrieval task, we compared our scheme with several typical baseline image retrieval methods built based on deep pooling features, including the average-pooled convolutional features under triplet architecture (Triplet), maximum activations of convolutions (MAC) \cite{azizpour2015generic}, regional MAC (R-MAC) and Generalized-Mean (GeM) pooling \cite{radenovic2018fine}. For a fair comparison, we evaluated all the methods mentioned above using the same CNN model and trained the deep features with triplet loss.    

\textbf{Deep Embeddings with Hard Generation.} Our proposed two-stage hard sample generation method is a novel generation method that enables the network to synthesize informative triplet more efficiently, leading to boosts in performance. The state-of-the-art deep embeddings with hard generation are included in this work for comparisons: the hard-negative generation method DAML \cite{duan2018deep}, HDML \cite{zheng2019hardness} and symmetrical synthesis (SS) \cite{gu2020symmetrical}.} 

\textbf{Comparison Analysis.} Table \ref{cub-result}, \ref{Cars196-result}, and \ref{online-result} show the quantitative results of our method and the comparison methods on the CUB-200-2011, Cars196, and SOP datasets. The feature dimension is 512 by default in the three tables above, except for special cases where the dimension is marked with superscript. Bold numbers indicate the best results and blue color represents the second-best performance. As we can see from the tables, our method achieves the best Recall@K and mAP on the CUB-200-2011, Cars196 datasets (Table \ref{cub-result} and Table \ref{Cars196-result}), while obtaining at least the second-best performance on the Stanford dataset (Table \ref{online-result}). {It should be noted that the image retrieval accuracy gain (Table \ref{online-result}) introduced by our method on the Stanford Online Products dataset is smaller than the improvement on the other two datasets (Table \ref{cub-result} and Table \ref{Cars196-result}). The possible reason may come from the fact that the Stanford Online Products dataset \cite{oh2016deep} consists of more diverse images than the other two datasets. The existing data already contains many hard samples and the left improving space for hard sample generation on Stanford Online Products dataset is limited.} For studying the performance of our method with varying embedding sizes, we conducted additional experiments on all three using a small embedding dimension. We can find that the performance of our proposed method can outperform the previous schemes even if the feature dimension is reduced to 128.

Compared with the image retrieval schemes with deep pooling features (MAC, R-MAC, GeM), the methods with hard sample generation (DAML, HDML, SS, and our scheme) produce better results. Compared with the DAML, HDML, and SS which improve performance by generating negative samples, our scheme can further boost the retrieval performance. In addition, our scheme can also obtain at least the second-best performance in the result of F1 and NMI. In summary, our proposed two-stage hard sample generation method is highly competitive in both retrieval and clustering tasks.

\subsection{Combine Hard Generation and Hard Mining}

The motivation of our hard sample generation approach is to synthesize difficult sample pairs from a large number of simple samples to obtain valid information. In contrast, existing hard sample mining strategies directly select the most informative sample pairs from real samples for training. To further explore whether the generated hard samples and real hard samples are complementary, we design to combine the generated samples with the mined samples for training.
We perform experiments to analyze the effect of combining our proposed method with a variety of hard sample mining strategies.

\textbf{Hard Sample Mining Methods.} Triplet loss needs to mine training tuples from the available mini-batch. In our study, all of our experiments are based on triplet architecture, so we choose the four most representative tuple mining strategies for our experiments. These strategies include random tuple mining (R) \cite{hu2014discriminative}, semihard triplet mining (Semi)\cite{schroff2015facenet}, softhard triplet mining (Soft) \cite{Yu_2018_ECCV} and distance-weighted tuple mining (D) \cite{wu2017sampling}. In this experiment, our proposed algorithm combined with random tuple mining fully equivalent to the two-stage hard sample generation method (THSG), which can be regarded as the baseline of our methods.

\begin{table}[!bp]
	\centering
    \begin{tabular}{lcccccccc}
	\toprule
    Method         & R@1  & R@2  & R@4  & R@8  & NMI  & F1   & MAP  \\   \midrule
    Triplet(R)         & 58.0          & 70.1          & 80.3          & 86.0          & 64.4          & 32.6          & 29.9          \\
    Triplet(Semi)       & 59.2          & 70.9          & 80.8          & 86.4          & 64.8          & 33.3          & 30.8          \\
    Triplet(Soft)       & 60.3          & 72.1          & 81.8          & 87.2          & 65.9          & 34.6          & 31.6          \\
    Triplet(D)       & 62.5          & 72.9          & 82.0          & 88.7          & 66.3          & 34.8          & 32.3          \\
    THSG   & 61.3          & 72.8          & 82.3          & 89.1          & 65.6          & 33.8          & 31.5          \\
    THSG(Semi) & 61.3          & 72.5          & 81.8          & 88.2          & 64.9          & 32.7          & 31.8          \\
    THSG(Soft) & 62.4          & 74.0          & 82.8          & 89.2          & 66.3          & 34.5          & 32.4          \\
    THSG(D) & \textbf{63.2} & \textbf{74.6} & \textbf{83.5} & \textbf{89.5} & \textbf{66.7} & \textbf{35.9} & \textbf{33.0} \\
	\bottomrule
    \end{tabular}
	\caption{Experimental results of THSG combining hard mining methods compared with their respective baselines on the CUB-200-2011 dataset(\%).}
	\label{Cub-mining-result}
\end{table}

\textbf{Implementation Details.}
In order to warrant unbiased comparability in Table \ref{Cub-mining-result}, \ref{Cars196-mining-result}, and \ref{online-mining-result}, our training protocol follows settings of \cite{roth2020revisiting}. The difference from the previous experimental setting is that we resize images to $224 \times 224$ and utilize a ResNet50 architecture with frozen Batch-Normalization layers and embedding dimensionality 128 .
When combining two different hard triplet for training, we only replace the original deep distance metric loss $\mathcal{L}_{org}$ with the hard mining deep distance metric loss $\mathcal{L}_{mining}$ and keep the other  training steps of ours method unchanged. Therefore, the final deep embedding learning (\ref{eq:final-function1}) is rewritten as:
\begin{align}
 \mathcal{L}_{F}&=  w_o\mathcal{L}_{mining} + w_l\mathcal{L}_{class} + w_h\mathcal{L}_{hard}
\label{eq:final-function3}
\end{align}

\textbf{Analysis of Results.}
We can find that experimental results in all evaluation criteria are further boosted by combining our method with a hard sample mining. Both in the retrieval task and in the clustering task, our method combined with hard sample mining results in better performance compared to the individual method. The performance improvement implies there is complementary information in the mined and generated hard samples. In particular, our generation method combined with simple mining strategies, such as random tuple mining or semihard triplet mining, yields an even more obvious improvement in results. Thus, our method does generate hard samples with information that is difficult to mine in the original data.

\begin{table}[ht]
	\centering
    \begin{tabular}{lcccccccc}
	\toprule
    Method         & R@1  & R@2  & R@4  & R@8  & NMI  & F1   & MAP  \\   \midrule
    Triplet(R)         & 67.8          & 78.2          & 85.8          & 91.2          & 60.1          & 27.2          & 26.9          \\
    Triplet(Semi)       & 70.6          & 80.1          & 86.7          & 91.5          & 61.3          & 29.7          & 28.8          \\
    Triplet(Soft)       & 76.9          & 84.6          & 90.3          & 94.1          & 62.8          & 30.6          & 31.3          \\
    Triplet(D)       & 77.7          & 85.7          & 90.8          & 94.0          & 64.4          & 33.1          & 31.9          \\
    THSG   & 80.2          & 87.2          & 91.7          & 95.0          & 66.2          & 34.8          & 31.9          \\
    THSG(Semi) & 79.8          & 86.4          & 91.5          & 95.0          & 66.0          & 35.4          & 34.2          \\
    THSG(Soft) & 80.3          & 87.0          & 91.8          & 95.2          & 66.4          & 35.6          & 34.2          \\
    THSG(D) & \textbf{81.6} & \textbf{88.4} & \textbf{92.5} & \textbf{95.6} & \textbf{67.5} & \textbf{36.7} & \textbf{34.6} \\
	\bottomrule
    \end{tabular}
	\caption{Experimental results of THSG combining hard mining methods compared with their respective baselines on the Cars196 dataset(\%).}
	\label{Cars196-mining-result}
\end{table}

\begin{table}[ht]
	\centering
    \begin{tabular}{lcccccccc}
	\toprule
    Method         & R@1  & R@10 & R@100 & NMI  & F1   & MAP  \\  \midrule
    Triplet(R)         & 71.0          & 85.3          & 93.6          & 88.8          & 30.9          & 38.7          \\
    Triplet(Semi)       & 76.2          & 88.8          & 95.4          & 89.9          & 36.2          & 44.3          \\
    Triplet(Soft)       & 76.3          & 89.1          & 95.6          & 89.8          & 35.8          & 44.1          \\
    Triplet(D)       & 77.8          & 90.2          & 95.9          & 90.1          & 36.9          & 45.8          \\
    THSG   & 76.2          & 88.7          & 95.2          & 89.6          & 34.7          & 43.4          \\
    THSG(Semi) & 78.0          & 89.9          & 95.8          & 90.1          & \textbf{37.3} & 45.9          \\
    THSG(Soft) & 77.7          & 89.8          & 95.7          & 90.0          & 36.6          & 45.5          \\
    THSG(D) & \textbf{78.5} & \textbf{90.3} & \textbf{95.9} & \textbf{90.1} & 37.1          & \textbf{46.3} \\
	\bottomrule
    \end{tabular}
	\caption{Experimental results of THSG combining hard mining methods compared with their respective baselines on the SOP dataset(\%).}
	\label{online-mining-result}
\end{table}

\subsection{Subjective Comparison}
\textbf{Visualized Retrieval Cases.} We selected several queries from the utilized dataset and depicted the retrieval results in Fig. \ref{fig:retrieval-vis}. The depicted methods are our work, the HDML of work {\cite{zheng2019hardness}} and the basic triplet scheme. The visualized results show that our method achieves better search accuracy. It should be noted that for some cases, such as the last two queries in Fig. \ref{fig:retrieval-vis}, almost all the methods have many error results. 

\begin{figure*}[bht]
  \centering
  \includegraphics[width=1.0\linewidth]{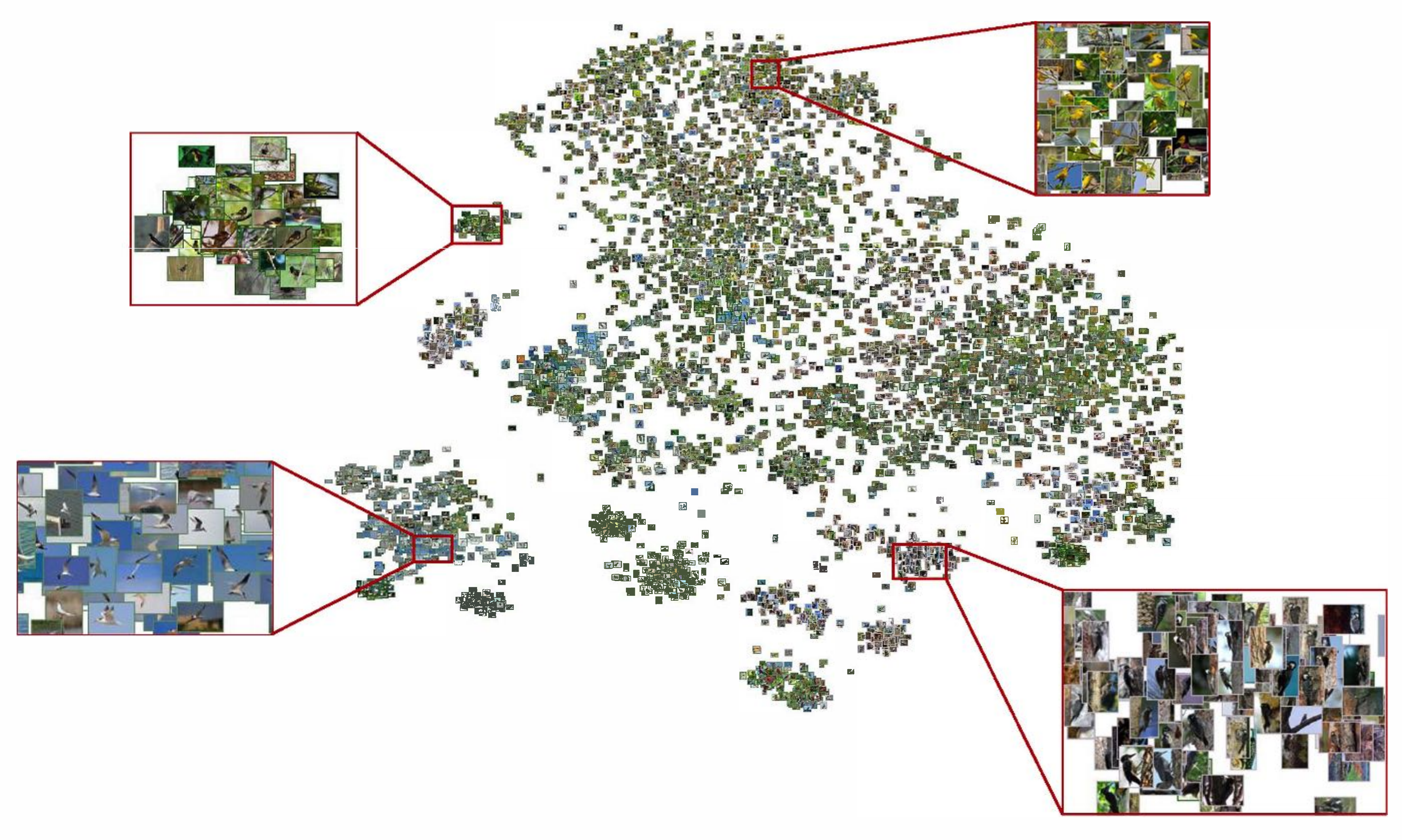}
  \caption{{Visualization of the proposed method with Barnes-Hut t-SNE \cite{van2014accelerating} on the CUB-200-2011 dataset, where the color of the border for each image represents the label. Four selected areas are magnified for a better view.}}
  \label{fig:cub200}
\end{figure*}

\begin{figure*}[bth]
  \centering
  \includegraphics[width=1.0\linewidth]{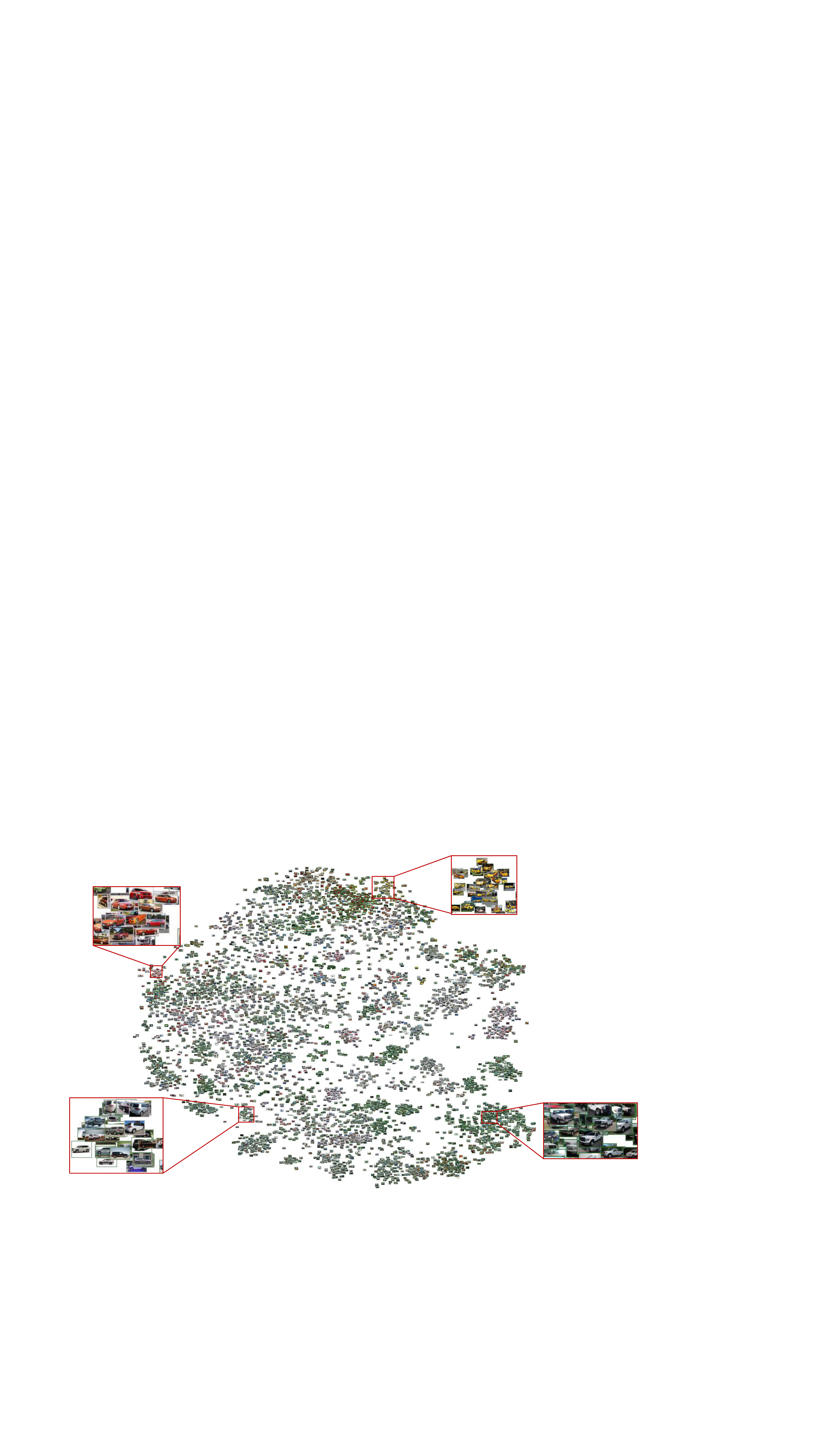}
  \caption{{Visualization of the proposed method with Barnes-Hut t-SNE \cite{van2014accelerating} on the Cars196 dataset, where the color of the border for each image represents the label. Four selected areas are magnified for a better view.}}
  \label{fig:cars}
\end{figure*}

\textbf{Embedding Visualization.} We visualized the learned embeddings for the CUB-200-2011 and Cars196 datasets by using t-SNE \cite{van2014accelerating}, as denoted by Fig. \ref{fig:cub200} and Fig. \ref{fig:cars}. In each figure, several selected specific regions are enlarged to highlight the representative classes for easy observation. In each specific region, although the images in the same class suffer from large variations in poses, colors, and backgrounds, our proposed method still can group similar objects. The visualization results denote our learned embeddings have strong representation ability and thus can achieve better retrieval performance in an intuitive manner.

\subsection{Ablation Study}
\begin{figure}[htbp]
  \centering
\begin{minipage}[t]{0.43\textwidth}
  \centering
  \includegraphics[width=1.0\textwidth]{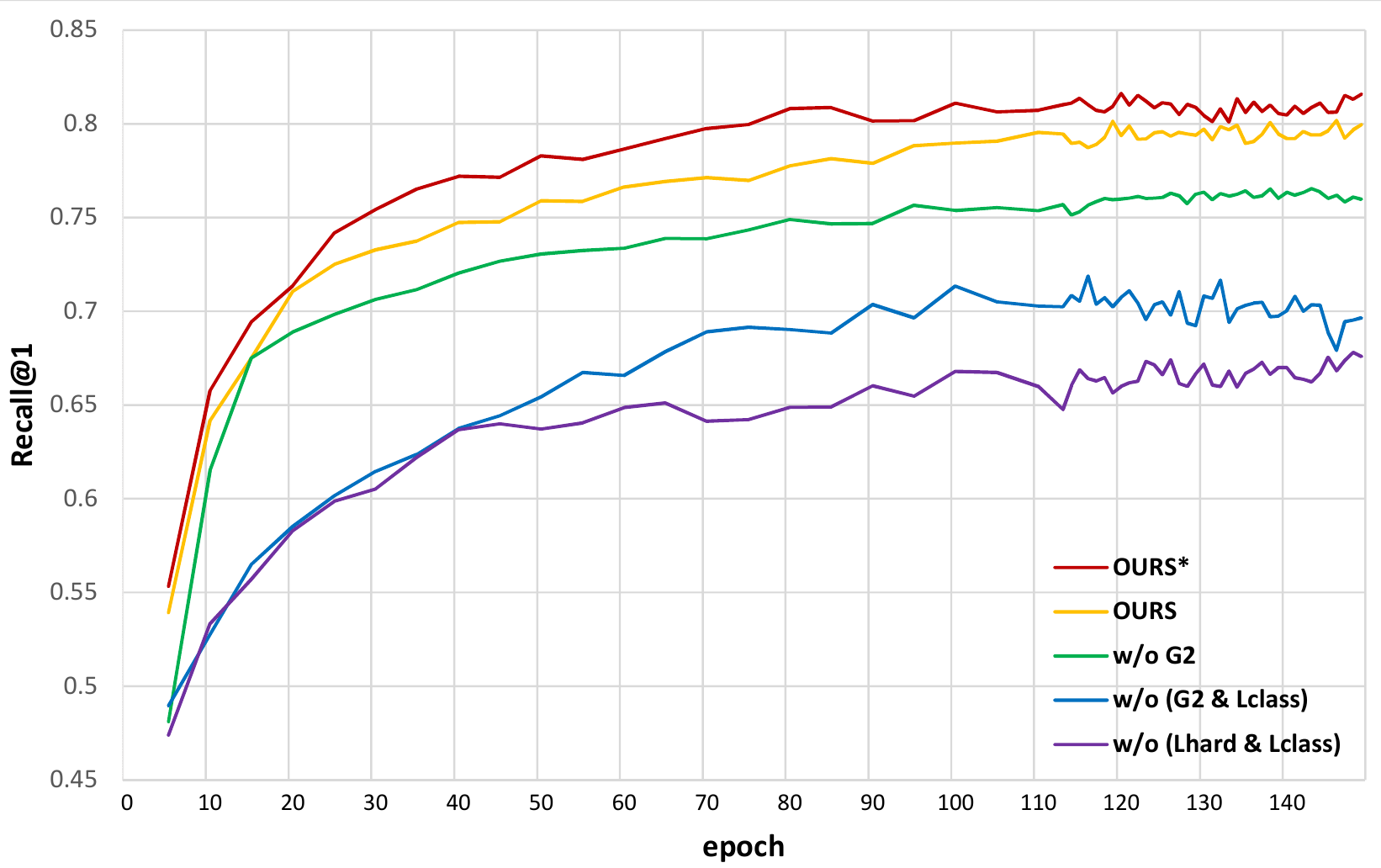}
  \caption{Performance comparisons of different settings during training.}
  \label{fig:Ablation_train}
  \end{minipage}
  \hspace{.15in}
\begin{minipage}[t]{0.43\textwidth}
   \centering
  \includegraphics[width=1.0\textwidth]{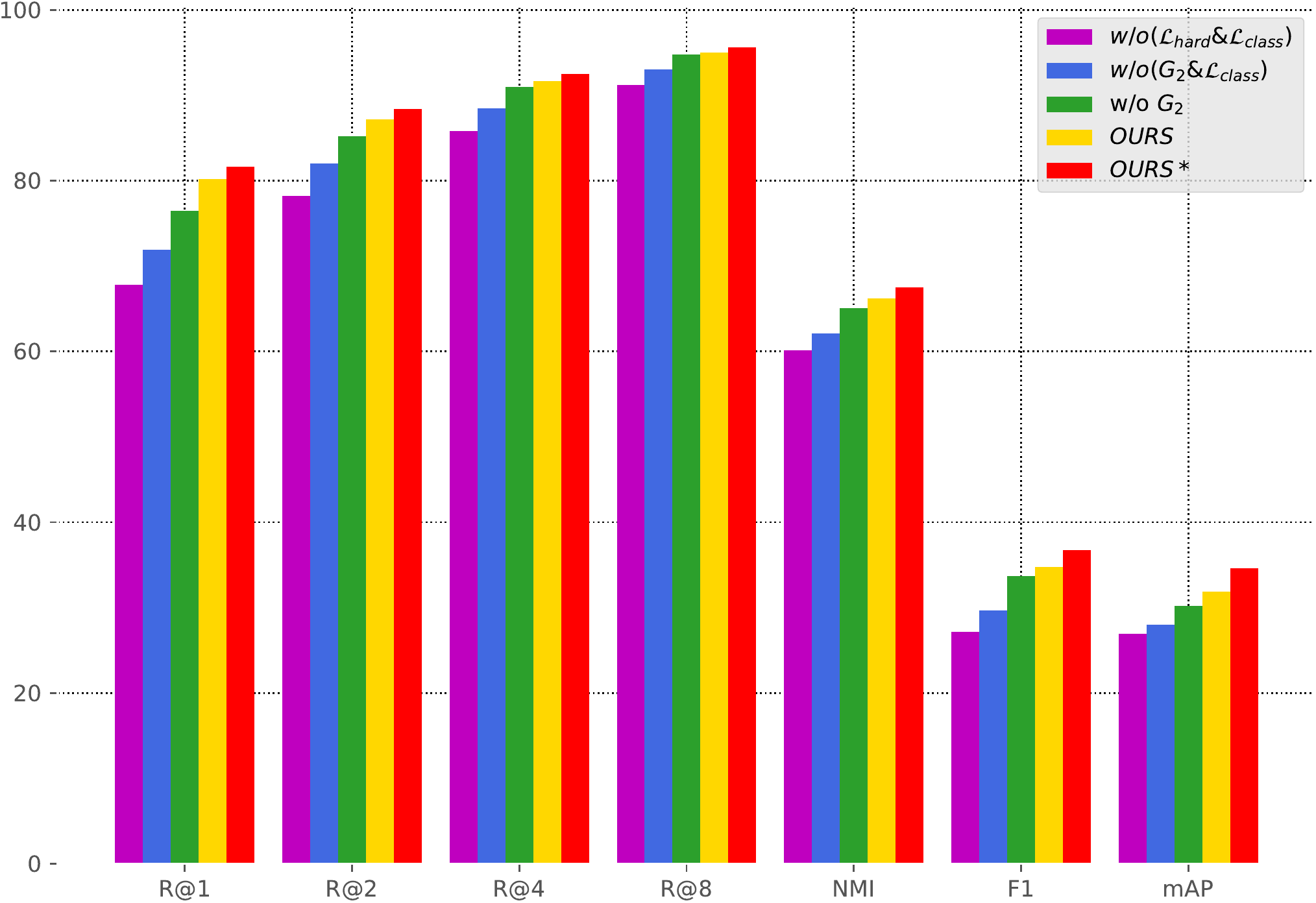}
  \caption{Ablation study: Result comparisons of different settings in Recall@Ks, NMI, F1 and mAP.}
  \label{fig:Ablation}
  \end{minipage}
\end{figure}

We conducted the ablation study of the proposed method on the Cars196 dataset. In Fig. \ref{fig:Ablation_train}, we show the results based on the Recall@1 criterion of different model settings, including our proposed method (OURS), the proposed method combined with distance-weighted hard mining (OURS$^*$), the proposed method without using the second stage of synthesis (w/o $G2$), the proposed method without using the second stage of synthesis and category loss $\mathcal{L}_{class}$ (w/o ($G2$ \& $\mathcal{L}_{class}$)), and the proposed method without both the generated hard sample loss $\mathcal{L}_{hard}$ and the category loss $\mathcal{L}_{class}$ (w/o ($\mathcal{L}_{hard}$ \& $\mathcal{L}_{class}$)), which is the baseline of using triplet architecture. 

We can find that the Recall@1 is further improved (80.2\% $\rightarrow$ 81.6\%) by combining our method with a hard sample mining. The performance improvement implies there is complementary information in the mined and generated hard samples. We remove the Generator $G_2$ and only use the first stage of generation to synthesize the anchor-positive pairs in hard triplets, then some performance loss is introduced (80.2\% $\rightarrow$ 76.5\%), which reveals the importance of hard sample generation which can extract the potential information from simple samples for embedding training. By further removing category loss $\mathcal{L}_{class}$, the performance suffers an obvious degradation again (76.5\% $\rightarrow$ 71.9\%), which tells that the category loss still can provide useful information under the metric learning architecture. Moreover, because we train the embedding model $F$ and the other hard sample generation models alternately, the absence of $\mathcal{L}_{class}$ will ruin the original obtained sample $\mathbf{X}$ and in turn impairs the generation of hard samples. Thus, the category loss $\mathcal{L}_{class}$ also plays a great role in our scheme. When the generated hard sample loss $\mathcal{L}_{hard}$ is totally removed (w/o $\mathcal{L}_{hard}$), we see the performance decreases a lot (71.9\% $\rightarrow$ 67.8\%), which verifies the effectiveness of the proposed stronger conditional synthesis scheme.  

To further verify the contributions of each component in our method, we perform the similar experiment on all quantitative criteria including Recall@K, NMI, F1, and mAP, which are depicted as Fig. \ref{fig:Ablation}. The quantitative results in Fig. \ref{fig:Ablation} also confirm the analysis as presented above, and the specific performance reduction brought to the model can be observed when each corresponding component is removed.

\section{Conclusion}
\label{sec:conclusion}
The existing mining-based triplet constructing methods directly learn the deep embeddings based on mining the hard samples, which will ignore the potential information of the easy samples. Although the recent hard sample generation schemes have tried to dig the potential information of the easy samples with the one-stage adversarial model, it is ignored the fact that the positives and negatives have different distributions and characteristics. In this study, we designed a two-stage hard sample generation scheme to achieve better embedding learning. 

We found that our proposed scheme achieves the best overall image retrieval and clustering performance on three large datasets (see Table \ref{cub-result}, \ref{Cars196-result}, and \ref{online-result}), indicating the effectiveness of the two-stage hard sample generation scheme. Specifically, our proposed method can conduct better results than the existing metric learning with hard sample generation. We also found that both the anchor-positive pairs generated by the conditional synthesis in the first stage and the negative samples generated in the second stage are crucial. Besides, by combining the generated hard samples and real hard samples selected by hard sample mining strategies, the performance can be further boosted (see Fig. \ref{fig:Ablation_train} and Fig. \ref{fig:Ablation}).

Our results provide compelling performance for image retrieval and clustering tasks through the proposed hard sample generation scheme.  In this paper, the usage of generated samples in our experiments is limited to the triplet metric learning architecture. In the future, we can consider exploring more efficient usage strategies for generated samples and we will conduct more hard sample generation research on the other loss architectures, such as the n-pair loss structure. 
In addition, there is a general problem in current hard sample generation schemes is that the generated samples are discarded after being used once. In the future, if schemes can be designed to efficiently reuse high-quality generated samples, it is possible to save computational resources and further accelerate network training.

\section{Acknowledgement}
{This work was supported in part by the 111 project (NO.B17007), Shandong Key Laboratory of Intelligent Buildings Technology (Grant No.SDIBT202006).}

\label{sec:ack}


\bibliography{mybibfile}

\begin{thebibliography}{10}
\expandafter\ifx\csname url\endcsname\relax
  \def\url#1{\texttt{#1}}\fi
\expandafter\ifx\csname urlprefix\endcsname\relax\def\urlprefix{URL }\fi
\expandafter\ifx\csname href\endcsname\relax
  \def\href#1#2{#2} \def\path#1{#1}\fi

\bibitem{li2015weakly}
Z.~Li, J.~Tang, Weakly supervised deep metric learning for
  community-contributed image retrieval, IEEE Transactions on Multimedia
  17~(11) (2015) 1989--1999.

\bibitem{girod2011mobile}
B.~Girod, V.~Chandrasekhar, D.~M. Chen, N.-M. Cheung, R.~Grzeszczuk, Y.~Reznik,
  G.~Takacs, S.~S. Tsai, R.~Vedantham, Mobile visual search, IEEE signal
  processing magazine 28~(4) (2011) 61--76.

\bibitem{zhu2019feature}
C.~Zhu, H.~Dong, S.~Zhang, Feature fusion for image retrieval with adaptive
  bitrate allocation and hard negative mining, IEEE Access 7 (2019)
  161858--161870.

\bibitem{lowe2004distinctive}
D.~G. Lowe, Distinctive image features from scale-invariant keypoints,
  International journal of computer vision 60~(2) (2004) 91--110.

\bibitem{perronnin2010large}
F.~Perronnin, Y.~Liu, J.~S{\'a}nchez, H.~Poirier, Large-scale image retrieval
  with compressed fisher vectors, in: 2010 IEEE Computer Society Conference on
  Computer Vision and Pattern Recognition, IEEE, 2010, pp. 3384--3391.

\bibitem{jegou2011aggregating}
H.~Jegou, F.~Perronnin, M.~Douze, J.~S{\'a}nchez, P.~Perez, C.~Schmid,
  Aggregating local image descriptors into compact codes, IEEE transactions on
  pattern analysis and machine intelligence 34~(9) (2011) 1704--1716.

\bibitem{azizpour2015generic}
H.~Azizpour, A.~Sharif~Razavian, J.~Sullivan, A.~Maki, S.~Carlsson, From
  generic to specific deep representations for visual recognition, in:
  Proceedings of the IEEE conference on computer vision and pattern recognition
  workshops, 2015, pp. 36--45.

\bibitem{babenko2015aggregating}
A.~Babenko, V.~Lempitsky, Aggregating local deep features for image retrieval,
  in: Proceedings of the IEEE international conference on computer vision,
  2015, pp. 1269--1277.

\bibitem{tolias2015particular}
G.~Tolias, R.~Sicre, H.~J{\'e}gou, Particular object retrieval with integral
  max-pooling of cnn activations, arXiv preprint arXiv:1511.05879.

\bibitem{arandjelovic2016netvlad}
R.~Arandjelovic, P.~Gronat, A.~Torii, T.~Pajdla, J.~Sivic, Netvlad: Cnn
  architecture for weakly supervised place recognition, in: Proceedings of the
  IEEE conference on computer vision and pattern recognition, 2016, pp.
  5297--5307.

\bibitem{radenovic2016cnn}
F.~Radenovi{\'c}, G.~Tolias, O.~Chum, Cnn image retrieval learns from bow:
  Unsupervised fine-tuning with hard examples, in: European conference on
  computer vision, Springer, 2016, pp. 3--20.

\bibitem{hu2014discriminative}
J.~Hu, J.~Lu, Y.-P. Tan, Discriminative deep metric learning for face
  verification in the wild, in: Proceedings of the IEEE conference on computer
  vision and pattern recognition, 2014, pp. 1875--1882.

\bibitem{hoffer2015deep}
E.~Hoffer, N.~Ailon, Deep metric learning using triplet network, in:
  International Workshop on Similarity-Based Pattern Recognition, Springer,
  2015, pp. 84--92.

\bibitem{wu2017sampling}
C.-Y. Wu, R.~Manmatha, A.~J. Smola, P.~Krahenbuhl, Sampling matters in deep
  embedding learning, in: Proceedings of the IEEE International Conference on
  Computer Vision, 2017, pp. 2840--2848.

\bibitem{yuan2017hard}
Y.~Yuan, K.~Yang, C.~Zhang, Hard-aware deeply cascaded embedding, in:
  Proceedings of the IEEE international conference on computer vision, 2017,
  pp. 814--823.

\bibitem{radenovic2018revisiting}
F.~Radenovi{\'c}, A.~Iscen, G.~Tolias, Y.~Avrithis, O.~Chum, Revisiting oxford
  and paris: Large-scale image retrieval benchmarking, in: Proceedings of the
  IEEE Conference on Computer Vision and Pattern Recognition, 2018, pp.
  5706--5715.

\bibitem{radenovic2018fine}
F.~Radenovi{\'c}, G.~Tolias, O.~Chum, Fine-tuning cnn image retrieval with no
  human annotation, IEEE transactions on pattern analysis and machine
  intelligence 41~(7) (2018) 1655--1668.

\bibitem{yu2018correcting}
B.~Yu, T.~Liu, M.~Gong, C.~Ding, D.~Tao, Correcting the triplet selection bias
  for triplet loss, in: Proceedings of the European Conference on Computer
  Vision (ECCV), 2018, pp. 71--87.

\bibitem{zhao2018adversarial}
Y.~Zhao, Z.~Jin, G.-j. Qi, H.~Lu, X.-s. Hua, An adversarial approach to hard
  triplet generation, in: Proceedings of the European Conference on Computer
  Vision (ECCV), 2018, pp. 501--517.

\bibitem{duan2018deep}
Y.~Duan, W.~Zheng, X.~Lin, J.~Lu, J.~Zhou, Deep adversarial metric learning,
  in: Proceedings of the IEEE Conference on Computer Vision and Pattern
  Recognition, 2018, pp. 2780--2789.

\bibitem{guo2019mode}
Y.~Guo, D.~An, X.~Qi, Z.~Luo, S.-T. Yau, X.~Gu, et~al., Mode collapse and
  regularity of optimal transportation maps, arXiv preprint arXiv:1902.02934.

\bibitem{cui2016fine}
Y.~Cui, F.~Zhou, Y.~Lin, S.~Belongie, Fine-grained categorization and dataset
  bootstrapping using deep metric learning with humans in the loop, in:
  Proceedings of the IEEE conference on computer vision and pattern
  recognition, 2016, pp. 1153--1162.

\bibitem{hermans2017defense}
A.~Hermans, L.~Beyer, B.~Leibe, In defense of the triplet loss for person
  re-identification, arXiv preprint arXiv:1703.07737.

\bibitem{Yu_2018_ECCV}
R.~Yu, Z.~Dou, S.~Bai, Z.~Zhang, Y.~Xu, X.~Bai, Hard-aware point-to-set deep
  metric for person re-identification, in: Proceedings of the European
  conference on computer vision (ECCV), 2018, pp. 188--204.

\bibitem{schroff2015facenet}
F.~Schroff, D.~Kalenichenko, J.~Philbin, Facenet: A unified embedding for face
  recognition and clustering, in: Proceedings of the IEEE conference on
  computer vision and pattern recognition, 2015, pp. 815--823.

\bibitem{zheng2019hardness}
W.~Zheng, Z.~Chen, J.~Lu, J.~Zhou, Hardness-aware deep metric learning, in:
  Proceedings of the IEEE Conference on Computer Vision and Pattern
  Recognition, 2019, pp. 72--81.

\bibitem{gu2020symmetrical}
G.~Gu, B.~Ko, Symmetrical synthesis for deep metric learning, arXiv preprint
  arXiv:2001.11658.

\bibitem{sohn2016improved}
K.~Sohn, Improved deep metric learning with multi-class n-pair loss objective,
  in: Advances in Neural Information Processing Systems, 2016, pp. 1857--1865.

\bibitem{chopra2005learning}
S.~Chopra, R.~Hadsell, Y.~LeCun, et~al., Learning a similarity metric
  discriminatively, with application to face verification, in: CVPR (1), 2005,
  pp. 539--546.

\bibitem{zhang2016understanding}
C.~Zhang, S.~Bengio, M.~Hardt, B.~Recht, O.~Vinyals, Understanding deep
  learning requires rethinking generalization, arXiv preprint arXiv:1611.03530.

\bibitem{mirza2014conditional}
M.~Mirza, S.~Osindero, Conditional generative adversarial nets, arXiv preprint
  arXiv:1411.1784.

\bibitem{oh2016deep}
H.~Oh~Song, Y.~Xiang, S.~Jegelka, S.~Savarese, Deep metric learning via lifted
  structured feature embedding, in: Proceedings of the IEEE Conference on
  Computer Vision and Pattern Recognition, 2016, pp. 4004--4012.

\bibitem{oh2017deep}
H.~Oh~Song, S.~Jegelka, V.~Rathod, K.~Murphy, Deep metric learning via facility
  location, in: Proceedings of the IEEE Conference on Computer Vision and
  Pattern Recognition, 2017, pp. 5382--5390.

\bibitem{wah2011caltech}
C.~Wah, S.~Branson, P.~Welinder, P.~Perona, S.~Belongie, The caltech-ucsd
  birds-200-2011 dataset.

\bibitem{krause20133d}
J.~Krause, M.~Stark, J.~Deng, L.~Fei-Fei, 3d object representations for
  fine-grained categorization, in: Proceedings of the IEEE International
  Conference on Computer Vision Workshops, 2013, pp. 554--561.

\bibitem{szegedy2015going}
C.~Szegedy, W.~Liu, Y.~Jia, P.~Sermanet, S.~Reed, D.~Anguelov, D.~Erhan,
  V.~Vanhoucke, A.~Rabinovich, Going deeper with convolutions, in: Proceedings
  of the IEEE conference on computer vision and pattern recognition, 2015, pp.
  1--9.

\bibitem{he2016deep}
K.~He, X.~Zhang, S.~Ren, J.~Sun, Deep residual learning for image recognition,
  in: Proceedings of the IEEE conference on computer vision and pattern
  recognition, 2016, pp. 770--778.

\bibitem{russakovsky2015imagenet}
O.~Russakovsky, J.~Deng, H.~Su, J.~Krause, S.~Satheesh, S.~Ma, Z.~Huang,
  A.~Karpathy, A.~Khosla, M.~Bernstein, et~al., Imagenet large scale visual
  recognition challenge, International journal of computer vision 115~(3)
  (2015) 211--252.

\bibitem{roth2020revisiting}
K.~Roth, T.~Milbich, S.~Sinha, P.~Gupta, B.~Ommer, J.~P. Cohen, Revisiting
  training strategies and generalization performance in deep metric learning,
  arXiv preprint arXiv:2002.08473.

\bibitem{van2014accelerating}
L.~Van Der~Maaten, Accelerating t-sne using tree-based algorithms, The Journal
  of Machine Learning Research 15~(1) (2014) 3221--3245.

\end{thebibliography}

\end{document}